\documentclass[10pt,twocolumn,letterpaper]{article}

\usepackage[pagenumbers]{cvpr} 

\usepackage{graphicx}
\usepackage{amsmath}
\usepackage{amssymb}
\usepackage{booktabs}

\usepackage{multirow}
\usepackage{xcolor}
\usepackage{array}
\usepackage{multirow}
\usepackage{mathtools}
\usepackage{overpic}
\usepackage{stmaryrd}

\usepackage[pagebackref,breaklinks,colorlinks]{hyperref}

\usepackage[capitalize]{cleveref}
\crefname{section}{Sec.}{Secs.}
\Crefname{section}{Section}{Sections}
\Crefname{table}{Table}{Tables}
\crefname{table}{Tab.}{Tabs.}
\crefname{figure}{Fig.}{Figs.}
\Crefname{figure}{Figure}{Figures}
\crefname{equation}{Eq.}{Eqs.}
\Crefname{equation}{Equation}{Equations}
\crefname{appendix}{Appx.}{Appxs.}
\Crefname{Appendix}{Appendix}{Appendices}

\def\cvprPaperID{****} 
\def\confName{CVPR}
\def\confYear{2023}

\makeatletter
\DeclareRobustCommand\onedot{\futurelet\@let@token\@onedot}
\def\@onedot{\ifx\@let@token.\else.\null\fi\xspace}

\def\eg{\emph{e.g}\onedot}

\def\ie{\emph{i.e}\onedot}

\def\vs{\emph{vs}\onedot}
\def\wrt{w.r.t\onedot}

\makeatother

\newcommand{\ours}{GraphSCNet}

\newcommand{\tight}[1]{\hspace{1pt}{#1}{\hspace{1pt}}}

\newcolumntype{P}[1]{>{\centering\arraybackslash}p{#1}}
\newlength{\wdth}

\newcommand{\ptitle}[1]{\noindent\textbf{#1}\hspace{5pt}}

\begin{document}

\title{Deep Graph-based Spatial Consistency for Robust Non-rigid\\Point Cloud Registration}

\author{
Zheng Qin$^{1}$\hspace{8pt}Hao Yu$^{2}$\hspace{8pt}Changjian Wang$^{1}$\hspace{8pt}Yuxing Peng$^{1}$\hspace{8pt}Kai Xu$^{1}$\thanks{Corresponding author: kevin.kai.xu@gmail.com.}\\
$^1$National University of Defense Technology\hspace{10pt}$^{2}$Technical University of Munich
}
\maketitle



\begin{abstract}

We study the problem of outlier correspondence pruning for non-rigid point cloud registration.
In rigid registration, spatial consistency has been a commonly used criterion to discriminate outliers from inliers. It measures the compatibility of two correspondences by the discrepancy between the respective distances in two point clouds.
However, spatial consistency no longer holds in non-rigid cases and outlier rejection for non-rigid registration has not been well studied.
In this work, we propose Graph-based Spatial Consistency Network (\emph{\ours{}}) to filter outliers for non-rigid registration.
Our method is based on the fact that non-rigid deformations are usually locally rigid, or local shape preserving.
We first design a local spatial consistency measure over the deformation graph of the point cloud, which evaluates the spatial compatibility only between the correspondences in the vicinity of a graph node.
An attention-based non-rigid correspondence embedding module is then devised to learn a robust representation of non-rigid correspondences from local spatial consistency.
Despite its simplicity, \ours{} effectively improves the quality of the putative correspondences and attains state-of-the-art performance on three challenging benchmarks.
Our code and models are available at \url{https://github.com/qinzheng93/GraphSCNet}.

\end{abstract}
\vspace{-10pt}


\section{Introduction}
\label{sec:intro}

Non-rigid point cloud registration is a fundamental and critical problem in computer graphics, computer vision, and robotics. It aims at recovering the non-rigid warping function that transforms a source point cloud to a target one.
In practice, the two point clouds are usually incomplete and share partial and even low overlap, which considerably increases the difficulty of registration.

Estimating the warping function relies on extracting accurate correspondences.
Benefiting from the recent advances in deep point representation~\cite{qi2017pointnet,wang2019dynamic,thomas2019kpconv,vaswani2017attention}, learning-based matching methods~\cite{deng2018ppfnet,gojcic2019perfect,choy2019fully,huang2021predator,li2022lepard,qin2022geometric,trappolini2021shape,wu2020pointpwc,puy2020flot} have obtained significantly high quality of putative correspondences. However, similar success has yet to be achieved in deformable cases. Under significant deformation, these methods are inevitably prone to outliers, which can drastically degrade the accuracy of registration.


\begin{figure}[t]
  \centering
  \begin{overpic}[width=1.0\linewidth]{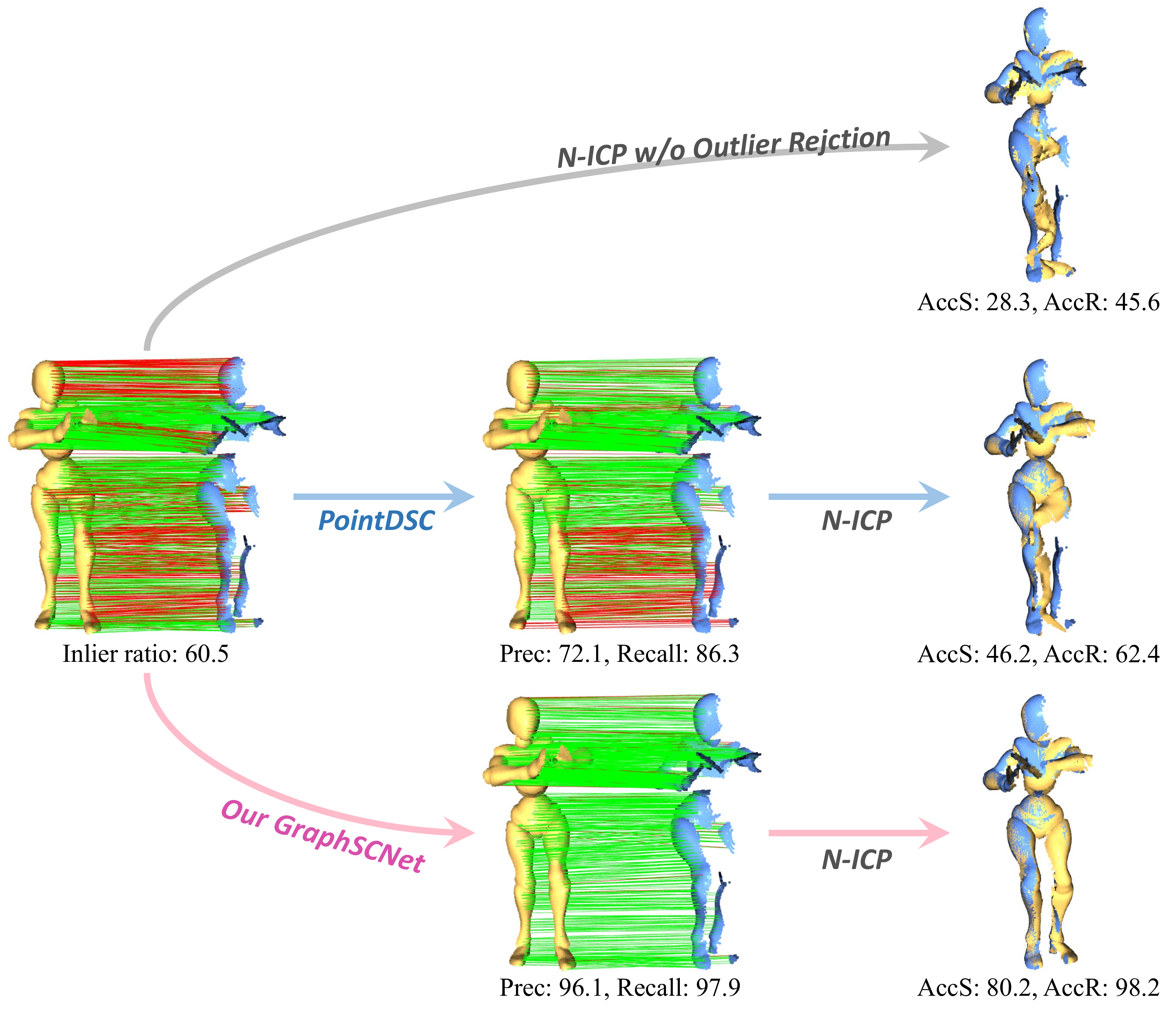}
  \end{overpic}
  \vspace{-21pt}
  \caption{\ours{} Overview. Given a set of putative correspondences for non-rigid registration, \ours{} can accurately prune the outliers among them while preserve the inliers, which contributes to significantly better registration results.}
  \label{fig:teaser}
  \vspace{-15pt}
\end{figure}

Outlier rejection is a common technique for robust point cloud registration.
However, most existing methods focus on rigid registration.
An effective method of outlier pruning for non-rigid registration has so far been missing.
On the one hand, a rigid transformation can be pinned down by a small set of inlier correspondences (\eg, a rotation can be determined by $3$ non-colinear inliers) such that sampling consensus methods (\eg, RANSAC~\cite{fischler1981random}) can effectively remove the outliers and recover the alignment transformation in a hypothesize-and-verify manner. However, non-rigid registration requires dense and thoroughly-distributed correspondences to precisely describe the deformation, thus preventing the application of sampling consensus methods.
On the other hand, rigid transformations preserve Euclidean distance between every pair of points. This spatial consistency provides a strong necessary condition for finding inlier correspondences and has been extensively adopted in rigid registration~\cite{bai2021pointdsc,leordeanu2005spectral,chen2022sc2,lee2021deep}.
Apparently, such spatial consistency does not hold for non-rigid cases.
These difficulties make outlier rejection for non-rigid registration a challenging problem.

We propose an outlier rejection network named \emph{Graph-based Spatial Consistency Network} (\ours{}) tailored for non-rigid registration. Our method is designed around the local rigidity of non-rigid deformations, \ie, non-rigid deformations are locally isometric such that the local shape of the point cloud is approximately preserved.
We first design a \emph{graph-based local spatial consistency} measure on the deformation graph~\cite{sumner2007embedded} built over the source point cloud.
It measures the geometric compatibility between the correspondences in the vicinity of a given graph node.
Based on this measure, we propose an attention-based \emph{graph-based correspondence embedding} module to extract \emph{spatial-consistency-aware features} for correspondences, which are further used for discriminative classification of inlier and outlier.
Thanks to the powerful local spatial consistency, our method can effectively prune outliers in putative correspondences while keeping as many inliers as possible. To our knowledge, our method is the first \emph{learning-based outlier rejection for non-rigid point cloud registration}.
Extensive experiments on three challenging benchmarks demonstrate clear superiority of our method.
In particular, \ours{} outperforms the recent state-of-the-art NDP~\cite{li2022non} by over $10\%$ on AccS and AccR for both high- and low-overlap scenarios on the 4DMatch benchmark~\cite{li2022lepard}.

Our main contributions include:
\begin{itemize}
	\vspace{-8pt}
	\item An outlier rejection network for non-rigid point cloud registration which is, to our knowledge, the first learning-based approach to outlier correspondence pruning for non-rigid scenarios.
	\vspace{-8pt}
	\item A graph-based local spatial consistency which measures the local geometric compatibility between correspondences within a local region.
	\vspace{-8pt}
	\item An attention-based correspondence embedding module which encodes the local spatial consistency for learning robust correspondence representation.
\end{itemize} 

\section{Related Work}
\label{sec:related}

\ptitle{Point cloud correspondence.}
Extracting accurate correspondences between point clouds plays a crucial role in computer vision and graphics tasks.
Detection-based methods first extract geometrically-discriminative keypoints and their descriptors, either with hand-crafted~\cite{johnson1999using,rusu2008aligning,rusu2009fast,tombari2010unique} or learning-based~\cite{zeng20173dmatch,deng2018ppfnet,deng2018ppf,gojcic2019perfect,choy2019fully,bai2020d3feat,huang2021predator,ao2021spinnet} descriptors, which are then matched as correspondences.
However, it is difficult to detect repeatable keypoints between point clouds, especially in low-overlap cases, such that detection-based methods still suffer from low inlier ratio.
Recently, detection-free methods~\cite{yu2021cofinet,qin2022geometric,li2022lepard} bypass keypoint detection by considering all possible point pairs in a coarse-to-fine matching pipeline, which significantly improves matching and registration accuracy.
There are also methods dedicated to non-rigid matching by explicitly modeling shape deformation~\cite{groueix20183d,trappolini2021shape,saleh2022bending} or leveraging functional maps~\cite{ovsjanikov2012functional,litany2017deep,donati2020deep}.
And scene flow estimation methods~\cite{liu2019flownet3d,wu2020pointpwc,puy2020flot,teed2021raft} predict the frame-to-frame motion of points in the scene.
Although great progress has been made, existing methods are still prone to outliers, which significantly harms the registration performance.

\ptitle{Non-rigid registration.}
To describe the non-rigid deformation, the warping function can be formulated into different representations, \eg, dense displacement field~\cite{li2022lepard}, dense affine transformation field~\cite{li2022non}, and embedded deformation graph~\cite{sumner2007embedded}.
Dense displacement field computes a 3D motion vector for each point in the scene, which is the most direct way to represent deformation.
Dense affine transformation field computes an affine transformation for each point, which can better model complex deformation.
Neural Deformation Pyramid~\cite{li2022non} establishes a hierarchical dense affine transformation field with multiple MLPs for coarse-to-fine non-rigid registration.
And embedded deformation graph~\cite{sumner2007embedded} parameterizes the deformation with a set of graph nodes connected with undirected edges, where each node is associated with an affine transformation.
This can be efficiently solved by the Non-rigid Iterative Closest Point (N-ICP) algorithm~\cite{li2008global}.
NNRT~\cite{bozic2020neural} proposes a differentiable N-ICP solver for end-to-end training, and~\cite{bozic2021neural} learns a deformation graph in a data-driven manner.
There are still other warping function formulations, and we refer the readers to~\cite{deng2022survey} for more details.

\ptitle{Outlier rejection for point cloud registration.}
Pruning outliers in rigid registration has been broadly studied. The most popular methods are RANSAC~\cite{fischler1981random} and its variants~\cite{chum2003locally,barath2018graph,barath2020magsac++}, which solve for the rigid transformation in a hypothesize-and-verify manner.
However, they suffer from slow convergence and could degenerate under high outlier ratio.
Other methods~\cite{leordeanu2005spectral,chen2022sc2} leverage spatial consistency to suppress outliers. Recent learning-based methods~\cite{choy2020deep,pais20203dregnet,bai2021pointdsc} filter outliers with a neural network.
PointDSC~\cite{bai2021pointdsc} designs a spatial consistency non-local module to prune outliers and attains promising rigid registration performance.
Nevertheless, due to complex deformations, similar success has yet to be achieved in non-rigid registration.
A closely related work to ours is~\cite{huang2008non}, which extends the traditional spectral matching~\cite{leordeanu2005spectral} technique to geodesic space. However, it computes pairwise geodesic distances between correspondences, which is time-consuming. And geodesic distance could be erroneous and unstable due to occlusion.
In this work, we fill this gap with \ours{} for efficient and accurate non-rigid outlier pruning.


\section{Method}
\label{sec:method}

\subsection{Overview}
\label{sec:overview}

Given a source point cloud $\mathcal{P} = \{\textbf{p}_i \in \mathbb{R}^3 \mid i = 1, ..., N\}$ and a target point cloud $\mathcal{Q} = \{\textbf{q}_i \in \mathbb{R}^3 \mid i = 1, ..., M\}$, non-rigid registration aims to recover the warping function $\mathcal{W}: \mathbb{R}^3 \rightarrow \mathbb{R}^3$ that transforms $\mathcal{P}$ to $\mathcal{Q}$.
To solve for the warping function, a set of correspondences $\mathcal{C} = \{(\mathbf{x}_i, \mathbf{y}_i) \in \mathbb{R}^6 \mid \mathbf{x}_i \in \mathcal{P}, \mathbf{y}_i \in \mathcal{Q} \}$ between two point clouds are first extracted.
Then the warping function $\mathcal{W}$ can be solved by minimizing the following cost function:
\begin{equation}
E = \lambda_c E_{\text{corr}} + \lambda_r E_{\text{reg}},
\label{eq:cost-function}
\end{equation}
where $E_{\text{corr}}$ is a correspondence term which minimizes the residuals of the correspondences after being warped, and $E_{\text{reg}}$ is a regularization term to encourage smoothness of deformations.
Nevertheless, the putative correspondences usually contain numerous outliers, which significantly harms the registration accuracy.
Due to complex deformations, it is difficult to filter outliers in non-rigid registration.
In this work, we first present the graph-based local spatial consistency which measures the compatibility of correspondences within a local region, and then propose an outlier rejection network for non-rigid registration.

\subsection{Graph-based Local Spatial Consistency}
\label{sec:local-spatial-consistency}


\begin{figure}[t]
  \centering
  \begin{overpic}[width=1.0\linewidth]{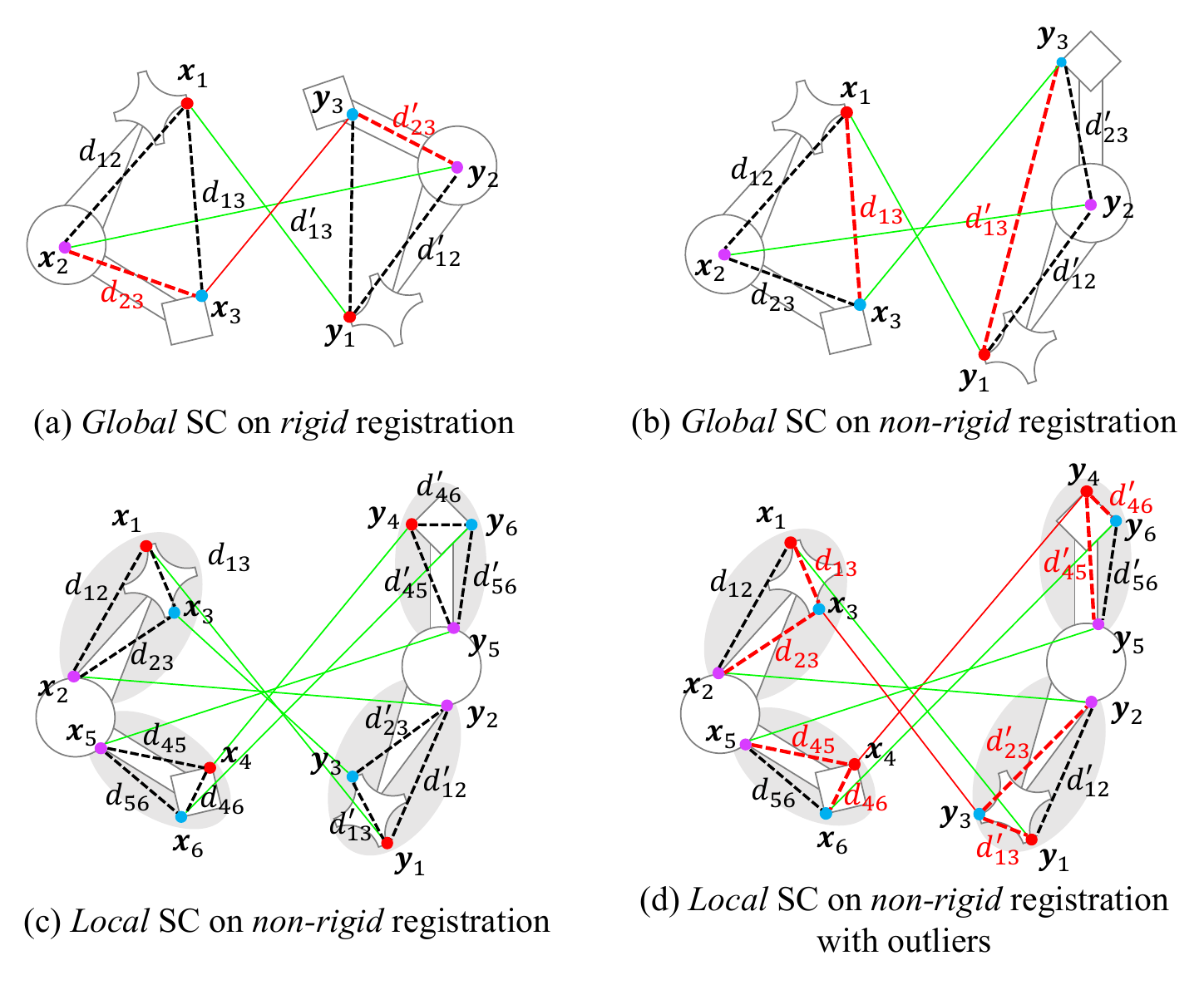}
  \end{overpic}
  \vspace{-20pt}
  \caption{Graph-based local spatial consistency for non-rigid registration. The {\color{green}green} lines represent the inliers while the outliers are in {\color{red}red}. And the inconsistent distances between two correspondences are also highlighted in {\color{red}red} dotted lines. (a) In rigid scenarios, the distances are identical between any two inliers, while being inconsistent if outliers exist. (b) In non-rigid scenarios, global spatial consistency does not hold as the distances between inliers could be different due to irregular movements. (c-d) Our graph-based local spatial consistency measures the distances between two correspondences within a local region based on local rigidity of deformations. }
  \label{fig:spatial-consistency}
  \vspace{-10pt}
\end{figure}

Spatial consistency is a widely used criterion~\cite{leordeanu2005spectral,bai2021pointdsc,chen2022sc2} to select inlier correspondences in rigid registration, \eg, length consistency which preserves the distance between every pair of points under arbitrary rigid transformations. Given two correspondences $c_i \tight{=} (\mathbf{x}_i, \mathbf{y}_i)$ and $c_j \tight{=} (\mathbf{x}_j, \mathbf{y}_j)$, the spatial consistency between them is computed as:
\begin{equation}
\theta^{*}_{i,j} = [1 - \frac{\delta_{i, j}^2}{\sigma_{d}^{2}}]_{+},
\label{eq:global-spatial-consistency}
\end{equation}
where $[\cdot]_{+} = \max(0, \cdot)$, $\delta_{i, j} = \big\lvert \lVert \mathbf{x}_i - \mathbf{x}_j \rVert - \lVert \mathbf{y}_i - \mathbf{y}_j \rVert \big\rvert$ is the difference between the respective distances in two point clouds, and $\sigma_d$ is a hyper-parameter to control the sensitivity to distance variation.
According to length consistency, $\delta_{i, j}$ should be small if they are both inliers, making $\theta^{*}_{i, j}$ close to $1$. But if there is at least one outlier, $\delta_{i, j}$ tends to be large due to the random distribution of the outliers, so $\theta^{*}_{i, j}$ should be $0$. See \cref{fig:spatial-consistency}(a) for a detailed illustration. This provides strong geometric support to reject outliers in rigid scenarios.

However, global spatial consistency no longer holds in non-rigid scenarios, especially between two inliers far from each other, as the points in different parts of the scene could follow inconsistent movements (see \cref{fig:spatial-consistency}(b)). But as noted in~\cite{igarashi2005rigid}, the local geometric shape is expected to be preserved and the warping function should be locally isometric and nearly rigid, \ie, local rigidity of deformations. Inspired by this insight, we propose to adopt spatial consistency in a local scope and devise a novel \emph{graph-based local spatial consistency}. 
Our method is based on the deformation graph~\cite{sumner2007embedded} built over the source point cloud.
We first sample a set of nodes $\mathcal{V} = \{\mathbf{v}_j \in \mathbb{R}^3 \mid j = 1, ..., V\}$ from $\mathcal{P}$ using \emph{uniform furthest point sampling}. We start from an arbitrary point in $\mathcal{P}$ and iteratively add the furthest point to the sampled nodes as a new node. The sampling process is repeated until the distances from all points in $\mathcal{P}$ to their nearest nodes are within $\sigma_n$.
Then, we assign each correspondence $c_i$ to its $k$-nearest nodes $\mathcal{N}_i$ according to the distances in $\mathcal{P}$.
Here $\mathcal{N}_i$ is constructed according to the Euclidean distance.
Given two points in a local region, their Euclidean distance is sufficiently consistent across two point clouds, but is more robust to occlusion than the geodesic distance.
The set of correspondences assigned to a node $\mathbf{v}_j$ is denoted as $\mathcal{C}_j = \{ c_i \mid \mathbf{v}_j \in \mathcal{N}_i \}$.
At last, our graph-based local spatial consistency is defined by computing \cref{eq:global-spatial-consistency} on the correspondence pairs assigned to a common node:
\begin{equation}
\theta_{i, j} = \begin{cases}
[1 - \delta_{i, j}^2 / \sigma_{d}^{2}]_{+}, & c_i \in \mathcal{C}_v \land c_j \in \mathcal{C}_v \\
0, & \text{otherwise}
\end{cases}.
\label{eq:spatial-consistency}
\end{equation}
Based on local rigidity, $\theta_{i, j}$ is expected to be close to $1$ if $c_i$ and $c_j$ are both inliers and be $0$ otherwise.
\cref{fig:spatial-consistency} compares our local spatial consistency with the global consistency.


\begin{figure*}[t]
  \centering
  \begin{overpic}[width=1.0\linewidth]{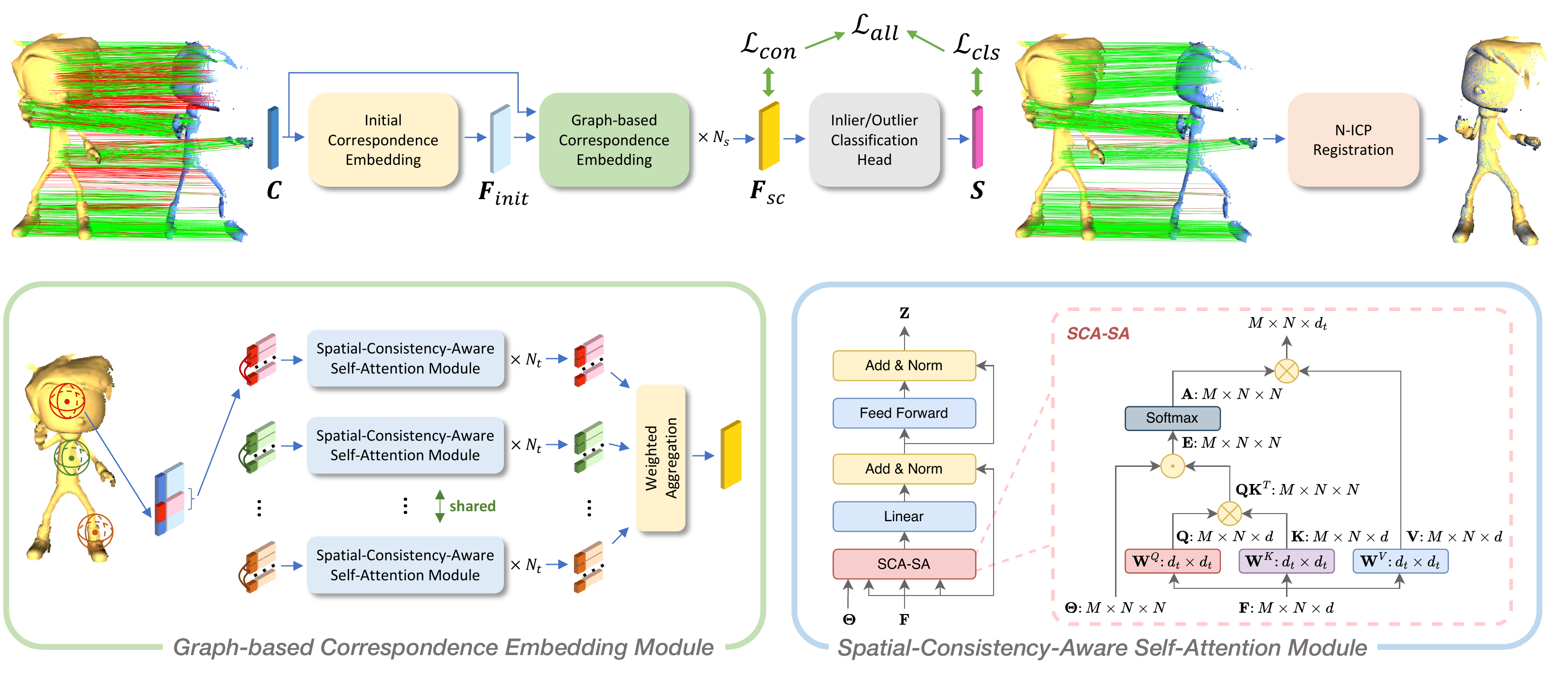}
  \end{overpic}
  \vspace{-20pt}
  \caption{Pipeline of \ours{}. Given a set of putative correspondences $\mathcal{C}$, our method first extracts initial features $\mathbf{F}_{\text{init}}$ from the point coordinates. The features are enhanced by a stack of graph-based non-rigid correspondence embedding module which encodes the local spatial consistency. The spatial-consistency-aware features $\mathbf{F}_{\text{sc}}$ are then used to predict the confidence scores $\mathbf{S}$. At last, N-ICP is used to estimate the warping function.
  }
  \label{fig:overview}
  \vspace{-15pt}
\end{figure*}

An alternative way to define local spatial consistency is to construct a $k$NN graph around each correspondence instead of the sampled nodes. However, this manner could have two main problems.
First, it requires more computation and memory usage to compute local spatial consistency around \emph{every} correspondence. This seriously restricts its scalability to large point clouds or dense correspondences.
Second, this fashion is sensitive to the density of putative correspondences. In practice, the distribution of correspondences could be extremely biased over the point cloud, and thus this manner is prone to be affected by the dense regions.
On the contrary, as our method is designed around uniformly sampled nodes, it has great advantage in efficiency and is naturally robust to density variation. Please refer to~\cref{sec:exp-ablation} for more detailed comparisons.

\subsection{Non-rigid Outlier Rejection Network}
\label{sec:outlier-rejection-network}

Based on the local spatial consistency, we then propose an attention-based \emph{Graph-based Spatial Consistency Network} (\emph{\ours{}}) for non-rigid outlier rejection. Given a set of putative correspondences, \ours{} leverages the graph-based local spatial consistency to remove the outliers from them.
The overall pipeline is illustrated in \cref{fig:overview}.

\ptitle{Initial feature embedding.}
For each input correspondence, we first concatenate the coordinates of the two endpoints into a $6$-d vector $\mathbf{c}_i = [\mathbf{x}_i; \mathbf{y}_i]$, which is then normalized to $\hat{\mathbf{c}}_i$ by subtracting the average over all correspondences.
Next, $\hat{\mathbf{c}}_i$ is transformed using Fourier positional encoding in~\cite{mildenhall2020nerf}.
As mentioned in~\cite{li2022non}, low-frequency encoding benefits fitting relatively rigid motion while high-frequency one can better model highly non-rigid motion. Recalling our goal to better capture local rigidity, we use relatively low frequency to encode the correspondences:
\begin{equation}
\mathbf{d}_i = [\hat{\mathbf{c}}; \hspace{5pt} \sin(2^{-1} \hat{\mathbf{c}}); \hspace{5pt} \cos(2^{-1} \hat{\mathbf{c}})] \in \mathbb{R}^{18}.
\end{equation}
At last, the encoded correspondence matrix $\mathbf{D} \in \mathbb{R}^{\lvert \mathcal{C} \rvert \times 18}$ is projected to a high-dimension feature matrix $\mathbf{F}_{\text{init}} \in \mathbb{R}^{\lvert \mathcal{C} \rvert \times d}$ by a shallow MLP, which is used as the initial correspondence embedding. And group normalization~\cite{wu2018group} and LeakyReLU are used after each layer in the MLP.

\ptitle{Graph-based correspondence embedding.}
With the initial correspondence embedding, we then design a \emph{Graph-based Correspondence Embedding Module} to enhance the feature representation of the correspondences with attention mechanism. The structure of this module is shown in \cref{fig:overview}~(bottom). Our method is based on the deformation graph constructed in \cref{sec:local-spatial-consistency} and consists of three steps.

First, we collect for each node $\mathbf{v}_j$ the correspondences in $\mathcal{C}_j$ and their associated features denoted as $\mathbf{F}_j \in \mathbb{R}^{\lvert \mathcal{C}_j \rvert \times d}$. Note that a correspondence could be assigned to more than one nodes and the nodes with $\mathcal{C}_j = \varnothing$ are ignored. We also collect the local spatial consistency of the correspondence pairs in $\mathcal{C}_j$, denoted as $\mathbf{\Theta}_j \in \mathbb{R}^{\lvert \mathcal{C}_j \rvert \times \lvert \mathcal{C}_j \rvert}$.

Next, we refine the features for the correspondences by a stack of \emph{Spatial-Consistency-Aware Self-Attention} (SCA-SA) module. Specifically, the feature matrix $\mathbf{F}_j$ is first projected into the query $\mathbf{Q}_j$, key $\mathbf{K}_j$ and value $\mathbf{V}_j$:
\begin{equation}
\mathbf{Q}_j = \mathbf{F}_j \mathbf{W}^Q, \hspace{10pt} \mathbf{K}_j = \mathbf{F}_j \mathbf{W}^K, \hspace{10pt} \mathbf{V}_j = \mathbf{F}_j \mathbf{W}^V,
\end{equation}
where $\mathbf{W}^Q$, $\mathbf{W}^K$, $\mathbf{W}^V \in \mathbb{R}^{d \times d}$ are the projection weights for query, key and value, respectively.
Inspired by~\cite{bai2021pointdsc}, we leverage the local spatial consistency to \emph{reweight} the attention scores in the original attention computation~\cite{vaswani2017attention}:
\begin{equation}
\mathbf{Z}'_j = \mathtt{LN}\Big(\mathbf{F}_j + \mathtt{MLP}\big( \mathtt{Softmax}(\mathbf{\Theta}_j \frac{\mathbf{Q}_j\mathbf{K}_j^T}{\sqrt{d}}) \mathbf{V}_j\big)\Big),
\end{equation}
where $\mathtt{LN}(\cdot)$ is layer normalization~\cite{ba2016layer}.
By injecting the graph-based local spatial consistency into self-attention, the correspondence pairs with strong spatial consistency are encouraged to have large attention scores, while the attention scores of the incompatible pairs are expected to be suppressed. This could push the outliers away from the inliers in the feature space, thus making the resultant features more discriminative.
The attention features are further projected by a two-layer feedforward network with residual connection to obtain the final output features:
\begin{equation}
\mathbf{Z}_j = \mathtt{LN}\big(\mathbf{Z}'_j + \mathtt{MLP}(\mathbf{Z}'_j)\big).
\end{equation}
\cref{fig:overview}~(bottom right) illustrates the structure and the computation graph of this module.

At last, for each correspondence, we consider its spatial compatibility \wrt different nodes and aggregate the features from all the nodes where it belongs as the final output features:
\begin{equation}
\mathbf{h}_i = \sum_{j \in \mathcal{N}_i} \alpha_{i, j} \mathbf{z}^j_i,
\end{equation}
where $\alpha_{i, j}$ is the skinning factor as in DynamicFusion~\cite{newcombe2015dynamicfusion}:
\begin{equation}
\alpha_{i, j} = \frac{\exp(-\lVert \mathbf{x}_i - \mathbf{v}_j \rVert^2 / (2 \sigma_n^2))}{\sum_{k \in \mathcal{N}_i} \exp(-\lVert \mathbf{x}_i - \mathbf{v}_k \rVert^2 / (2 \sigma_n^2))}.
\label{eq:skinning-factor}
\end{equation}
In non-rigid scenarios, it is unreliable to predict whether one correspondence is inlier or not from merely a single local area as there could be large deformation in it. On the contrary, our method considers all neighboring regions, which could improve the robustness of the extracted features.

\ptitle{Classification head.}
Given the spatial-consistency-aware features $\mathbf{F}_{\text{sc}} \in \mathbb{R}^{\lvert \mathcal{C} \rvert \times d}$ of the correspondences, we further adopt a three-layer MLP to predict the confidence score $s_i$ being an inlier for each correspondence. Group normalization~\cite{wu2018group} and LeakyReLU are used after the first two layers in the MLP, and sigmoid activation is applied after the last layer. The correspondences whose confidence scores are above a certain threshold $\tau_{s}$ are selected as inliers and the others are removed as outliers.

\subsection{Deformation Estimation}
\label{sec:deformation-estimation}

After obtaining the pruned correspondences, an embedded deformation graph~\cite{sumner2007embedded} is computed as the final warping function.
We first construct a deformation graph $\hat{\mathcal{G}} = \{\hat{\mathcal{V}}, \hat{\mathcal{E}}\}$ with a set of graph nodes $\hat{\mathcal{V}}$ and undirected edges $\hat{\mathcal{E}}$ connecting them.
The nodes are sampled from $\mathcal{P}$ as described in \cref{sec:local-spatial-consistency} with a distance threshold of $\sigma_g$.
Each point in $\mathcal{P}$ are assigned to its $k_g$ nearest nodes and two nodes are connected by an edge if there exists a point assigned to both of them.
$\mathcal{W}$ can then be approximated by a collection of local \emph{rigid} transformations $\{ (\mathbf{R}_j, \mathbf{t}_j) \}$ associated with each node $\hat{\mathbf{v}}_j$:
\begin{equation}
\mathcal{W}(\mathbf{p}_i) = \sum_{j \in \mathcal{N}_i} \alpha_{i,j} \big(\mathbf{R}_j (\mathbf{p}_i - \hat{\mathbf{v}}_j) + \mathbf{t}_j + \hat{\mathbf{v}}_j\big),
\label{eq:embedded-deformation}
\end{equation}
where $\alpha_{i, j}$ is computed as in \cref{eq:skinning-factor}.
Our final optimization objective is shown as in \cref{eq:cost-function}, where the correspondence term is the mean squared distance between the correspondences and an as-rigid-as-possible~\cite{igarashi2005rigid} regularization term is applied to constrain the smoothness of deformations:
\begin{equation}
\begin{aligned}
E_{\text{corr}} & = \sum_{(\mathbf{x}_i, \mathbf{y}_i) \in \mathcal{C}} \lVert \mathcal{W}(\mathbf{x}_i) - \mathbf{y}_i \rVert_{2}^{2} \\
E_{\text{reg}} & = \sum_{(\mathbf{v}_i, \mathbf{v}_j) \in \mathcal{E}} \lVert \mathbf{R}_{i}(\mathbf{v}_{j} - \mathbf{v}_{i}) + \mathbf{v}_{i} + \mathbf{t}_{i} - (\mathbf{v}_{j} + \mathbf{t}_{j}) \rVert_{2}^{2}
\end{aligned}.
\end{equation}
This problem can be efficiently solved by Non-rigid ICP (N-ICP) algorithm~\cite{li2008global,sumner2007embedded}.
Note that although embedded deformation is used, \ours{} is agnostic to deformation models and thus can facilitate any correspondence-based non-rigid registration methods.

\subsection{Loss Functions}

Our model is trained with two types of loss functions, including a classification loss and a consistency loss. The overall loss function is computed as $\mathcal{L}_{\text{all}} = \mathcal{L}_{\text{cls}} + \lambda \mathcal{L}_{\text{con}}$.

\ptitle{Classification loss.}
We formulate the prediction of the confidence scores of the correspondences as a binary classification problem. As inliers and outliers are usually very imbalanced in the putative correspondences, we supervise the confidence scores with a binary focal loss~\cite{lin2017focal}. The label of each correspondence $c_i = (\mathbf{x}_i, \mathbf{y}_i)$ is computed as:
\begin{equation}
s^{*}_i = \begin{cases}
1, & \lVert \mathcal{W}^{*}(\mathbf{x}_{i}) - \mathbf{y}_{i} \rVert < \tau_{d} \\
0, & \text{otherwise}
\end{cases},
\end{equation}
where $\mathcal{W}^{*}$ is the ground-truth deformation. And the classification loss is computed as:
\begin{equation}
\mathcal{L}_{\text{cls}} = - s^{*}_i (1 - s_i)^{\gamma} \log(s_i) - (1 - s^{*}_i) s_i^{\gamma} \log(1 - s_i),
\end{equation}
where $\gamma = 2$ is the focusing hyper-parameter as in \cite{lin2017focal}.

\ptitle{Consistency loss.}
Inspired by PointDSC~\cite{bai2021pointdsc}, we further adopt an auxiliary feature consistency loss so that the inliers are close to each other in the feature space and are far away from the outliers. However, due to the complexity of non-rigid deformations, feature consistency could not hold between two distant inlier correspondences. For this reason, we propose to supervise the feature consistency in each local region. 
For two correspondences $c_x, c_y \in \mathcal{C}_j$ of node $\textbf{v}_j$, we first compute their feature consistency as:
\begin{equation}
\delta_{x, y} = [1 - \frac{\lVert \hat{\mathbf{h}}_x - \hat{\mathbf{h}}_y \rVert^2}{\sigma_f^2}]_{+},
\end{equation}
where $\hat{\mathbf{h}}_x$ and $\hat{\mathbf{h}}_y$ are the correspondence features which are normalized onto a unit hyper-sphere, and $\sigma_f$ is a learnable tolerance parameter. The consistency loss is computed as:
\begin{equation}
\mathcal{L}_{\text{con}} = \frac{1}{\lvert \mathcal{V} \rvert^2} \sum_{\textbf{v}_j \in \mathcal{V}} \frac{1}{\lvert \mathcal{C}_j \rvert^2} \sum_{\textbf{c}_x \in \mathcal{C}_j} \sum_{\textbf{c}_y \in \mathcal{C}_j} \in \lVert \delta_{x, y} - \delta^{*}_{x, y} \rVert,
\end{equation}
where the ground-truth targets $\delta^{*}_{x, y} = 1$ if $c_x$ and $c_y$ are both inliers and $\delta^{*}_{x, y} = 0$ otherwise.

\section{Experiments}

We evaluate the efficacy of \ours{} on three challenging benchmarks: 4DMatch~\cite{li2022lepard} (\cref{sec:exp-4dmatch}), CAPE~\cite{ma2020learning,pons2017clothcap} (\cref{sec:exp-cape}) and DeepDeform~\cite{bozic2020deepdeform} (\cref{sec:exp-deepdeform}). Extensive ablation studies are also provided to better understand our design choices (\cref{sec:exp-ablation}). More implementation details and network settings are introduced in the appendix.

\ptitle{Metrics.}
Following~\cite{li2022lepard,li2022non}, we mainly evaluate $4$ metrics in the experiments:
(1) \emph{3D End Point Error} (EPE), the average errors over all warped points under the estimated and the ground-truth warp functions, (2) \emph{3D Accuracy Strict} (AccS), the fraction of points whose EPEs are below $2.5\text{cm}$ or relative errors are below $2.5\%$, (3) \emph{3D Accuracy Relaxed} (AccR), the fraction of points whose EPEs are below $5\text{cm}$ or relative errors are below $5\%$, and (4) \emph{Outlier Ratio} (OR), the fraction of points whose relative errors are above $30\%$.

\subsection{Evaluations on 4DMatch}
\label{sec:exp-4dmatch}

\ptitle{Dataset.}
4DMatch~\cite{li2022lepard} is a challenging synthetic benchmark for non-rigid point cloud registration, which is constructed using the animation sequences from DeformingThings4D~\cite{li20214dcomplete}. It consists of $1232$ sequences for training, $176$ for validation and $353$ for testing. The point cloud pairs in the testing sequences are divided into 4DMatch and 4DLoMatch based on a overlapping ratio threshold of $45\%$.
We use the preprocessed data from NDP~\cite{li2022non} which removes the testing pairs with nearly-rigid movements to better evaluate the performance on non-rigid scenarios.

\ptitle{Comparisons with state-of-the-art methods.}
\label{sec:exp-sota}
We first compare \ours{} to previous state-of-the-art non-rigid registration and scene flow estimation methods: NSFP~\cite{li2021neural}, Nerfies~\cite{park2021nerfies}, PointPWC-Net~\cite{wu2020pointpwc}, FLOT~\cite{puy2020flot}, DGFM~\cite{donati2020deep}, SyNoRiM~\cite{huang2022multiway}, and NDP~\cite{li2022non}. To evaluate the generality of our method, we adopt two recent deep correspondence extractors in the experiments, Lepard~\cite{li2022lepard} and GeoTransformer~\cite{qin2022geometric}. As shown in \cref{table:results-4dmatch}, our method outperforms the baselines by a large margin on both benchmarks, indicating the effectiveness of \ours{}.
On the two most important metrics \emph{AccS} and \emph{AccR}, our method significantly surpasses the previous best NDP by $11$ percentage points (pp) on 4DMatch and $14$ pp on 4DLoMatch.
Note that benefiting from the high-quality correspondences, our method achieves the new state-of-the-art results simply with N-ICP and achieves $10$ times acceleration than NDP ($0.2$s \vs $2$s).


\begin{table}[!t]
\scriptsize
\setlength{\tabcolsep}{2.5pt}
\centering
\begin{tabular}{l|cccc|cccc}
\toprule
\multirow{2}{*}{Model} & \multicolumn{4}{c|}{4DMatch} & \multicolumn{4}{c}{4DLoMatch} \\
 & EPE & AccS & AccR & OR & EPE & AccS & AccR & OR \\
\midrule
NSFP~\cite{li2021neural} & 0.265 & 8.7 & 18.7 & 65.0 & 0.495 & 0.4 & 1.6 & 84.8 \\
Nerfies~\cite{park2021nerfies} & 0.280 & 12.7 & 25.4 & 58.9 & 0.498 & 1.1 & 3.0 & 82.2 \\
PointPWC-Net~\cite{wu2020pointpwc} & 0.182 & 6.3 & 21.5 & 52.1 & 0.279 & 1.7 & 8.2 & 55.7 \\
FLOT~\cite{puy2020flot} & 0.133 & 7.7 & 27.2 & 40.5 & 0.210 & 2.7 & 13.1 & 42.5 \\
DGFM~\cite{donati2020deep} & 0.152 & 12.3 & 32.6 & 37.9 & 0.148 & 1.9 & 6.5 & 64.6 \\
SyNoRiM~\cite{huang2022multiway} & 0.099 & 22.9 & 49.9 & 26.0 & 0.170 & 10.6 & 30.2 & 31.1 \\
NDP~\cite{li2022non} & 0.077 & 61.3 & 74.1 & 17.3 & 0.177 & 26.6 & 41.1 & 33.8 \\
\midrule
\ours{} (\emph{ours})~+~\cite{li2022lepard} & \textbf{0.042} & \underline{70.1} & \underline{83.8} & \textbf{9.2} & \underline{0.102} & \underline{40.0} & \textbf{59.1} & \textbf{17.5} \\
\ours{} (\emph{ours})~+~\cite{qin2022geometric} & \underline{0.043} & \textbf{72.3} & \textbf{84.4} & \underline{9.4} & \underline{0.121} & \textbf{41.0} & \underline{58.3} & \underline{21.0} \\
\bottomrule
\end{tabular}
\vspace{-5pt}
\caption{
Comparisons with previous state-of-the-art methods on 4DMatch and 4DLoMatch.
\textbf{Boldfaced} numbers highlight the best and the second best are \underline{underlined}.
}
\label{table:results-4dmatch}
\vspace{-10pt}
\end{table}


\begin{table}[!t]
\scriptsize
\setlength{\tabcolsep}{3pt}
\centering
\begin{tabular}{l|cccc|cccc}
\toprule
\multirow{2}{*}{Model} & \multicolumn{4}{c|}{4DMatch} & \multicolumn{4}{c}{4DLoMatch} \\
 & Prec & Recall & AccS & AccR & Prec & Recall & AccS & AccR \\
\midrule
\multicolumn{9}{c}{\emph{Lepard}~\cite{li2022lepard}} \\
\midrule
w/o outlier rejection & 78.3 & 100.0 & 54.2 & 67.8 & 49.5 & 100.0 & 17.4 & 29.9 \\
VFC~\cite{ma2014robust} & 83.6 & \underline{93.2} & 63.6 & 76.4 & 54.6 & \underline{84.1} & 26.2 & 40.3 \\
PointCN~\cite{pais20203dregnet} & 87.0 & 89.0 & 63.2 & 78.1 & 71.8 & 75.6 & 31.6 & 50.7 \\
PointDSC~\cite{bai2021pointdsc} & \underline{88.7} & 92.2 & \underline{66.3} & \underline{80.3} & \underline{74.5} & 80.3 & \underline{35.2} & \underline{53.8} \\
\ours{} (\emph{ours}) & \textbf{93.0} & \textbf{95.7} & \textbf{70.1} & \textbf{83.8} & \textbf{83.0} & \textbf{88.6} & \textbf{40.0} & \textbf{59.1} \\
\emph{oracle} & 100.0 & 100.0 & 74.7 & 87.5 & 100.0 & 100.0 & 48.9 & 68.9 \\
\midrule
\multicolumn{9}{c}{\emph{GeoTransformer}~\cite{qin2022geometric}} \\
\midrule
w/o outlier rejection & 81.0 & 100.0 & 65.5 & 79.8 & 61.0 & 100.0 & 31.4 & 49.4 \\
VFC~\cite{ma2014robust} & 83.0 & \underline{96.0} & 67.1 & 79.6 & 63.2 & \textbf{91.6} & 33.8 & 50.5 \\
PointCN~\cite{pais20203dregnet} & 84.8 & 92.0 & 67.1 & 81.0 & 70.1 & 79.0 & 35.0 & 53.3 \\
PointDSC~\cite{bai2021pointdsc} & \underline{88.0} & 93.9 & \underline{69.2} & \underline{82.2} & \underline{73.7} & 81.8 & \underline{37.7} & \underline{55.0} \\
\ours{} (\emph{ours}) & \textbf{92.2} & \textbf{96.9} & \textbf{72.3} & \textbf{84.4} & \textbf{82.6} & \underline{86.8} & \textbf{41.0} & \textbf{58.3} \\
\emph{oracle} & 100.0 & 100.0 & 77.4 & 87.6 & 100.0 & 100.0 & 49.3 & 66.3 \\
\bottomrule
\end{tabular}
\vspace{-5pt}
\caption{
Comparisons with outlier rejection baselines on 4DMatch and 4DLoMatch.
\textbf{Boldfaced} numbers highlight the best and the second best are \underline{underlined}.
}
\label{table:results-rejection}
\vspace{-15pt}
\end{table}


\begin{figure*}[t]
  \centering
  \begin{overpic}[width=1.0\linewidth]{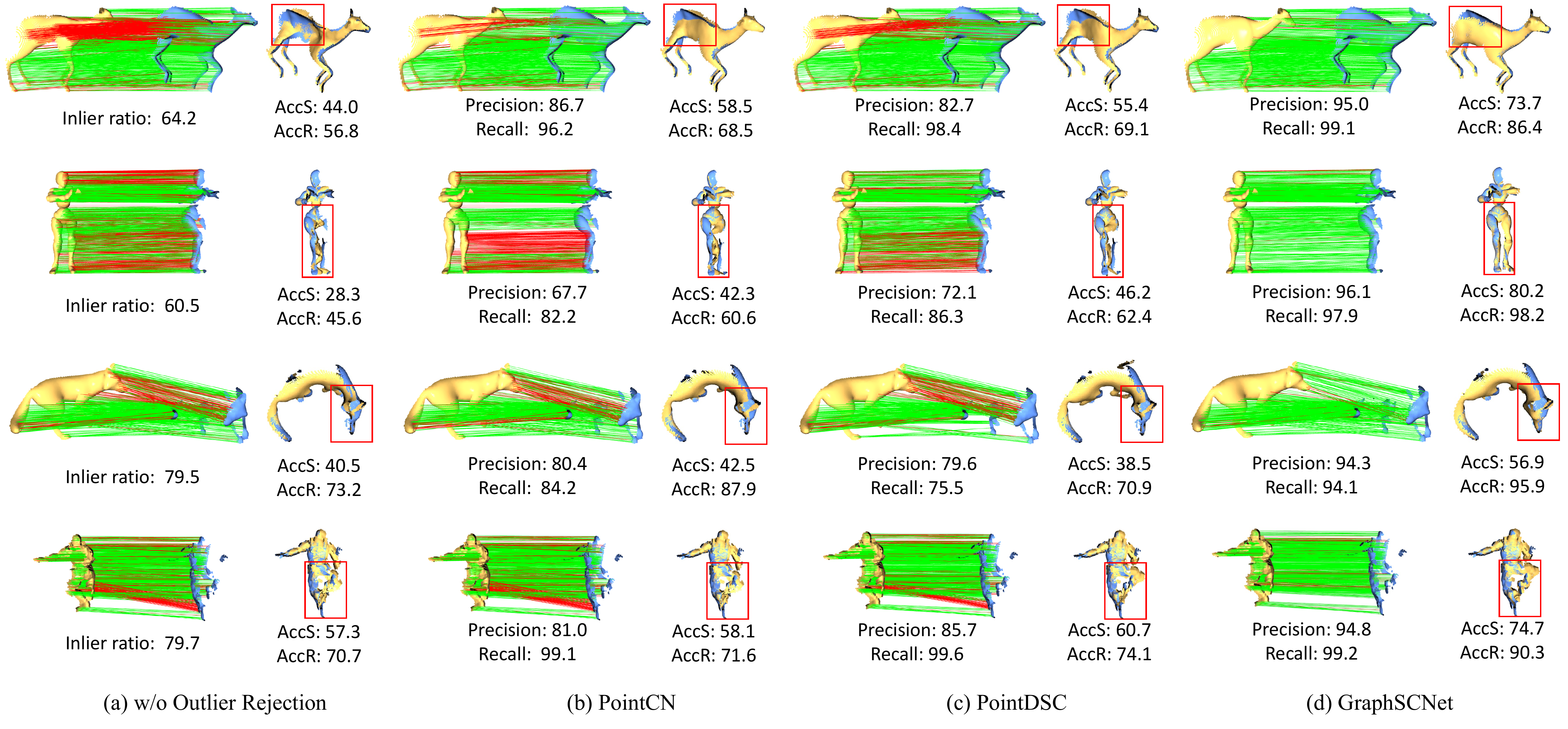}
  \end{overpic}
  \vspace{-20pt}
  \caption{
  Comparison of different methods on 4DMatch and 4DLoMatch.
  Our method provides much better outlier rejection results in low-overlap and large-deformation scenes and achives better registration results.
  Benefiting from the more accurate correspondences, our method successfully recover the geometry in non-overlap regions (see the registration results enclosed by the red boxes).}
  \label{fig:gallery-4dmatch}
  \vspace{-15pt}
\end{figure*}

\ptitle{Comparisons with outlier rejection methods.}
\label{sec:exp-rejection}
We compare to one traditional outlier rejection method, VFC~\cite{ma2014robust}, and two recent learning-based methods for \emph{rigid} registration, PointCN from 3DRegNet~\cite{pais20203dregnet} and PointDSC~\cite{bai2021pointdsc}, to evaluate the efficacy of our method. We also report the precision and recall of the predicted inliers to compare the inlier classification performance. For fair comparison, we adopt similar network macro-architecture for all the models and use the same configurations in N-ICP. For PointDSC, $2048$ correspondences are randomly sampled to avoid too huge memory footprint. We show the results on two correspondence extractors (Lepard and GeoTransformer) to compare the generality of the methods. And the results using the ground-truth inliers are also reported as \emph{oracle}. As shown in \cref{table:results-rejection}, the models with outlier rejection significantly surpass the models that do not prune outliers. And our method outperforms PointCN and PointDSC by a large margin on both benchmarks and attains very close results to the oracle, demonstrating the strong effectiveness of our design.
Note that our method attains both better precision and recall, especially in low-overlap scenarios, which means it \emph{rejects more outliers while preserving more inliers}.
This guarantees more thoroughly-distributed correspondences, facilitating more accurate non-rigid registration.

\ptitle{Qualitative results.}
\cref{fig:gallery-4dmatch} visualizes the correspondences and the registration results of different methods. Compared with the baselines, \ours{} prunes outliers more accurately while preserves more inliers, especially in low-overlap or large-deformation scenarios. And our method performs quite well in the scenes with symmetry (see the $2^{\text{nd}}$ row) or complex geometry (see the $4^{\text{th}}$ row). As there is little interference from outliers, our method successfully recover the geometry in non-overlap regions (see the registration results enclosed by the red box).

\subsection{Generalization from 4DMatch to CAPE}
\label{sec:exp-cape}

\ptitle{Dataset.}
CAPE~\cite{ma2020learning,pons2017clothcap} contains the complete scans of dynamic clothed humans. It consists of $15$ human subjects and provides accurate 3D mesh registrations.
We use the data preprocessed by~\cite{huang2022multiway} where each point coud contains $8192$ points.
To better study the performance on large deformations, we first align each point cloud pair with a rigid transformation by solving a mean least square problem~\cite{besl1992method}, and ignore the pairs whose mean residuals are below $10$cm.
At last, we obtain $11288$ point cloud pairs for evaluation.


\begin{table}[!t]
\scriptsize
\centering
\begin{tabular}{l|cccccc}
\toprule
Model & Prec & Recall & EPE & AccS & AccR & OR \\
\midrule
\multicolumn{7}{c}{\emph{Generalization to CAPE}} \\
\midrule
w/o outlier rejection & 38.0 & 100.0 & 0.143 & 20.9 & 41.3 & 68.2 \\
PointCN~\cite{pais20203dregnet} & 43.9 & \underline{65.0} & 0.132 & 28.3 & 50.3 & 63.6 \\
PointDSC~\cite{bai2021pointdsc} & \underline{55.8} & 63.2 & \underline{0.122} & \underline{34.4} & \underline{56.1} & \underline{60.5} \\
\ours{} (\emph{ours}) & \textbf{69.3} & \textbf{83.6} & \textbf{0.090} & \textbf{47.5} & \textbf{67.1} & \textbf{50.1} \\
\midrule
\multicolumn{7}{c}{\emph{Generalization to DeepDeform}} \\
\midrule
w/o outlier rejection & 43.3 & 100.0 & 0.146 & 17.8 & 37.0 & 64.0 \\
PointCN~\cite{pais20203dregnet} & 47.1 & \textbf{68.0} & 0.149 & 18.9 & 38.3 & 63.9 \\
PointDSC~\cite{bai2021pointdsc} & \underline{54.1} & \underline{65.1} & \underline{0.139} & \underline{21.7} & \underline{42.1} & \underline{61.4} \\
\ours{} (\emph{ours}) & \textbf{60.7} & 64.2 & \textbf{0.134} & \textbf{24.1} & \textbf{44.2} & \textbf{59.1} \\
\bottomrule
\end{tabular}
\vspace{-5pt}
\caption{
Generalization results from 4DMatch to DeepDeform and CAPE.
\textbf{Boldfaced} numbers highlight the best and the second best are \underline{underlined}.
}
\label{table:results-deepdeform-cape}
\vspace{-15pt}
\end{table}

\ptitle{Quantitative results.}
We investigate the generality of our method on CAPE.
To this end, we train all the models on 4DMatch and directly evaluate the models on CAPE without fine-tuning.
The input correspondences are extracted with GeoTransformer~\cite{qin2022geometric} which is also trained 4DMatch.
As shown in \cref{table:results-deepdeform-cape}(top), \ours{} achieves significant improvements over the baseline methods. Our method surpasses the second best PointDSC by over $13$ pp on precision and AccS, $20$ pp on recall, and $11$ pp on AccR. Note that our method not only achieves better precision, but very high recall, indicating that it prunes more outliers while preserves more inliers. 
As the human pose variations in CAPE are relatively large, the baseline methods fail to effectively distinguish inlier and outlier correspondences.
Nevertheless, our method is still effective and has strong robustness thanks to the local spatial consistency.
Please refer to the appendix for more detailed qualitative results.

\subsection{Generalization from 4DMatch to DeepDeform}
\label{sec:exp-deepdeform}

\ptitle{Dataset.}
DeepDeform~\cite{bozic2020deepdeform} consists of real-world partial RGB-D images scanned by a RGB-D camera. It contains $400$ scenes with over $390$K RGB-D frames.
We project the depth images into point clouds and leverage the dense scene flow annotations to construct the point cloud pairs.
And we adopt the same preprocessing as in~\cref{sec:exp-cape} with a threshold of $5$cm to remove the nearly-rigid pairs.
As a result, we obtain $1011$ point cloud pairs for evaluation.

\ptitle{Quantitative results.}
Following~\cref{sec:exp-cape}, we train all the models on 4DMatch and directly test them on DeepDeform without fine-tuning to invesigate the generality of our method to real-world scenarios. And GeoTransformer trained 4DMatch is adopted to generate the initial correspondences.
As shown in \cref{table:results-deepdeform-cape}(bottom), PointCN achieves only marginal improvements over the model without outlier rejection. As it determines outliers based on only the coordinates of each single correspondence without considering the geometry of point clouds, its generality is unsatisfactory. PointDSC obtains considerably better results than PointCN benefiting from the global spatial consistency. However, it still lacks the capability to handle non-rigid deformations. On the contrary, our method outperforms the three baseline methods by a large margin as it leverages local rigidity to remove outliers, which better models the deformations. These results have demonstated the strong transferability and generality of our method to unseen domains.

\subsection{Ablation Studies}
\label{sec:exp-ablation}

We further conduct extensive ablation studies to provide a better understanding of the design choices in \ours{}. In the following experiments, we use GeoTransformer as the prior correspondence extractor.


\begin{table}[!t]
\scriptsize
\setlength{\tabcolsep}{3pt}
\centering
\begin{tabular}{l|cccc|cccc}
\toprule
\multirow{2}{*}{Model} & \multicolumn{4}{c|}{4DMatch} & \multicolumn{4}{c}{4DLoMatch} \\
 & Prec & Recall & AccS & AccR & Prec & Recall & AccS & AccR \\
\midrule
(a.1) $\sigma_n=0.04$ & 91.4 & 96.4 & 70.6 & 83.5 & 79.3 & 85.2 & 38.4 & 56.5 \\
(a.2) $\sigma_n=0.08$* & \underline{92.2} & \textbf{96.9} & \textbf{72.3} & \textbf{84.4} & \underline{82.6} & \underline{86.8} & 41.0 & 58.3 \\
(a.3) $\sigma_n=0.12$ & \textbf{92.7} & 96.4 & \underline{72.0} & \underline{84.2} & \textbf{82.9} & 86.4 & \textbf{41.8} & \textbf{58.9} \\
(a.4) $\sigma_n=0.16$ & 92.1 & \underline{96.8} & 71.7 & 84.0 & 82.4 & \textbf{87.1} & \underline{41.7} & \underline{58.8} \\
(a.5) $\sigma_n=0.32$ & 89.2 & 95.4 & 69.5 & 82.4 & 78.6 & 83.7 & 39.3 & 56.6 \\
\midrule
(b.1) w/ FPS* & \textbf{92.2} & 96.9 & 72.3 & \textbf{84.4} & \textbf{82.6} & \textbf{86.8} & \textbf{41.0} & \textbf{58.3} \\
(b.2) w/ RS & 92.1 & \textbf{97.0} & \textbf{72.5} & \textbf{84.4} & 82.2 & 86.1 & \textbf{41.0} & 58.1 \\
\midrule
(c.1) Nodes w/ FPS* & \textbf{92.2} & \textbf{96.9} & \textbf{72.}3 & \textbf{84.4} & \textbf{82.6} & \textbf{86.8} & \textbf{41.0} & \textbf{58.3} \\
(c.2) $k$NN graph & 91.4 & 96.5 & 71.9 & 84.0 & 79.5 & 85.3 & 40.1 & 57.4 \\
\midrule
(d.1) $k=1$ & 89.1 & 96.6 & 59.7 & 69.4 & 75.6 & 85.9 & 23.6 & 31.4 \\
(d.2) $k=3$ & \underline{92.0} & \textbf{97.1} & \underline{71.1} & \underline{84.0} & \underline{81.6} & \underline{86.6} & 38.3 & 55.5 \\
(d.3) $k=6$* & \textbf{92.2} & \underline{96.9} & \textbf{72.3} & \textbf{84.4} & \textbf{82.6} & \textbf{86.8} & \textbf{41.0} & \textbf{58.3} \\
(d.4) $k=9$ & \textbf{92.2} & 96.6 & 69.8 & 83.0 & \textbf{82.6} & 85.8 & \underline{38.6} & \underline{56.5} \\
\bottomrule
\end{tabular}
\vspace{-5pt}
\caption{
Ablation studies on 4DMatch and 4DLoMatch.
Asterisk (*) indicates the default settings in our method.
\textbf{FPS}: furthest point sampling.
\textbf{RS}: random sampling.
\textbf{Boldfaced} numbers highlight the best and the second best are \underline{underlined}.
}
\label{table:results-ablation}
\vspace{-15pt}
\end{table}

\ptitle{Node sampling.}
We first study the influence of the distribution of nodes. First, we vary the distance threshold $\sigma_n$ from $0.04$ to $0.32$ for node sampling. Note that we do not change the node sampling settings in N-ICP.
As shown in \cref{table:results-ablation}(a), our method attains similar performance under different $\sigma_n$, and the performance get worse if $\sigma_n$ is too small or too large. If $\sigma_n$ is too small, each local region is limited so that there could not be enough context. But if $\sigma_n$ is too large, the local spatial consistency could be broken.

Next, we replace the uniform furthest point sampling with uniform random sampling in \cref{table:results-ablation}(b), and two sampling methods achieve comparable results. The model with furthest point sampling performs slightly better as it generates more stably distributed nodes. These results prove the strong robustness of our method to the distribution of nodes.

\ptitle{Graph construction.}
We further study the influence of the graph structure to compute local spatial consistency. First, we build a $k$NN graph which connects each correspondence to its $k$ nearest correspondences as described in \cref{sec:local-spatial-consistency}, where $k=32$ to fit the GPU memory and the same network architecture is used for outlier rejection. From \cref{table:results-ablation}(c), our method surpasses this counterpart on all the metrics, especially in low-overlap cases. We argue that our advantage is two-fold. First, each correspondence can only be connected to limited neighbors in the $k$NN graph due to high memory usage, which fails to provide enough geometric context. Second, the learned features are predominated by the spatial areas with high densities in the $k$NN graph such that the representation ability is degraded.

Next, we vary the number of nodes $k$ where each correspondence is assigned when building the graph in \cref{table:results-ablation}(d). The results are significantly degraded when $k=1$. One the one hand, considering only one local region for each correspondence harms the robustness of the learned features to large deformation. On the other hand, the geometric context in a local region is insufficient when $k$ is too small. And the performance also gets worse with a large $k$. The distances between correspondences and nodes could be too far such that the local spatial consistency is broken in this case.

\section{Conclusion}

We have proposed \ours{} to effectively prune outlier correspondences for non-rigid point cloud registration. 
Based on the local rigidity of deformations, we introduce a graph-based local spatial consistency criterion to measure the compatibility of two correspondences.
Next, we further design a non-rigid correspondence embedding module which leverages the local spatial consistency to extract correspondence features.
The spatial-consistency-aware correspondence features are further used to filter outliers.
To our knowledge, this is the first learning-based outlier rejection method for non-rigid registration.
Extensive experiments on three benchmarks have demonstrated the efficacy of our method.
However, as our method is based on deformation graphs and local rigidity, it could have difficulty in modeling sudden geometric changes.
In the future, we would like to extend our method for end-to-end non-rigid registration.

\ptitle{Acknowledgement.}
This work is supported in part by the NSFC (62132021), the National Key R\&D Program of China (2018AAA0102200, 2021ZD0140408), and the Logistics Major Research Plan (145BHQ090003000X03).

\appendix

This supplementary material provides the implementation details of \ours{} and the baselines~(\cref{appx:implementation-details}), the details of the metrics~(\cref{appx:metrics}), more experiments and analysis~(\cref{appx:additional-experiments}), the details of N-ICP to estimate the warping function~(\cref{appx:deformation-estimation}), and discusses the limitations of our method~(\cref{appx:limitations}).

\section{Implementation Details}
\label{appx:implementation-details}

\ptitle{Network architecture.}
In the initial feature embedding, we use a three-layer MLP with ($256$, $256$, $256$) channels to project the correspondence embedding to a high-dimension representation. Group normalization~\cite{wu2018group} and LeakyReLU are used after each layer in the MLP.

Unless otherwise noted, we use $3$ correspondence embedding modules to generate the spatial-consistency-aware features, while each contains $2$ SCA-SA modules. All layers in the models have $d=256$ feature channels.
The node coverage is $\sigma_n = 0.08\text{m}$. For each correspondence, we use $k=6$ neighboring nodes to construct the graph. And the distance tolerance when computing spatial consistency is $\sigma_{d} = 0.08\text{m}$. 

At last, we adopt another three-layer MLP with ($128$, $64$, $1$) channels to classify each correspondences. Group normalization~\cite{wu2018group} and LeakyReLU are used after the first two layers in the MLP, and sigmoid activation is applied after the last layer. We select the correspondences whose confidence scores are above $\tau_{s} = 0.4$ as inliers and the others are removed as outliers.

\begin{figure*}[t]
  \centering
  \begin{overpic}[width=0.95\linewidth]{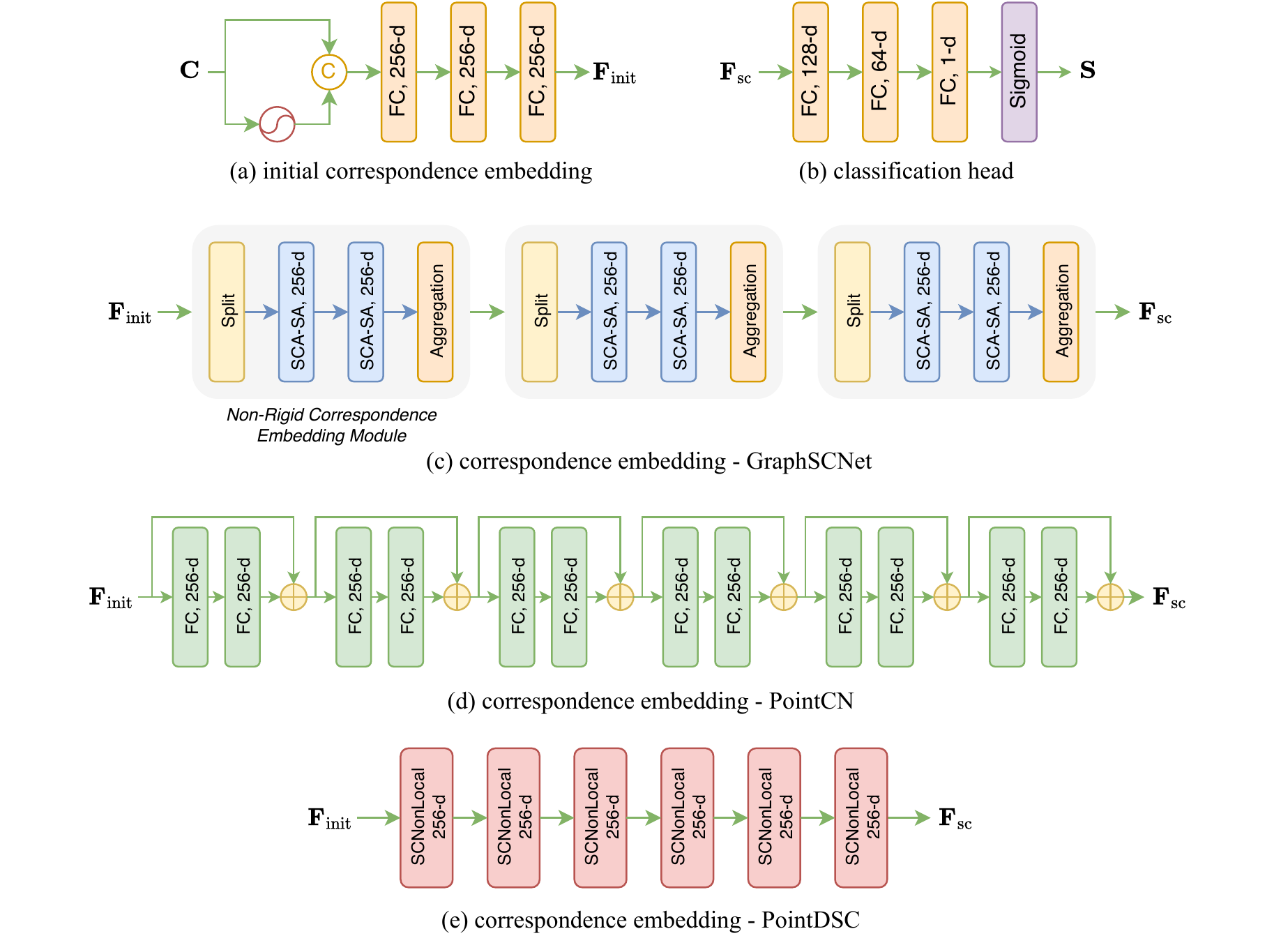}
  \end{overpic}
  \caption{
  Network architecture.
  }
  \label{fig:architecture}
\end{figure*}

\ptitle{Baselines.}
For the baseline models PointCN~\cite{pais20203dregnet} and PointDSC~\cite{bai2021pointdsc}, the initial feature embedding and the classification head are the same as aforementioned. In PointCN, we replace the correspondence embedding modules with $6$ MLP blocks, each of which consists of two linear layers with residual connection. In PointDSC, we use $6$ SCNonLocal~\cite{bai2021pointdsc} modules to learn the correspondence features. Due to memory limit, we randomly sample $2048$ input correspondences in PointDSC. The architectures of different models are illustrated in \cref{fig:architecture}. All the layers in the baseline models have $256$ feature channels as in \ours{}.

\ptitle{Training and testing.}
We implement and evaluate our method with PyTorch~\cite{paszke2019pytorch} on an NVIDIA $2080$Ti GPU.
The models are trained with Adam optimizer~\cite{kingma2014adam} for $40$ epochs. The batch size is $1$ and the weight decay is $10^{-6}$. The learning rate starts from $10^{-4}$ and decays exponentially by $0.05$ after each epoch. During training, we regard the correspondences as inliers if their residuals under the ground-truth deformation are below $\tau = 0.04\text{m}$, and outliers otherwise.
For data augmentation, we adopt a relatively weak data augmentation as in~\cite{yew2022regtr} with a random rotation sampled from $[0, 10^{\circ}]$ and a random translation sampled from $\mathcal{N}(0, 0.05)$.

In the experiments on 4DMatch, as the training data has been used to train the correspondence extractor, the putative correspondences on the training set are almost all inliers. In this case, the training data cannot provide effective supervision to train an outlier rejection network. To solve this problem, we split the official validation sequences by $90\%$/$10\%$ for training/validation, respectively, and evaluate the models on the official testing squences.


\section{Metrics}
\label{appx:metrics}

Following~\cite{li2022non}, we mainly evaluate our method using $4$ metrics: 3D End Point Error, 3D Accuracy Strict, 3D Accuracy Relaxed and Outlier Ratio.

\emph{3D End Point Error} (EPE) measures the average error over all warped points under the estimated and the ground-truth warping functions $\mathcal{W}(\cdot)$ and $\mathcal{W}^{*}(\cdot)$:
\begin{equation}
\mathtt{EPE} = \frac{1}{\lvert \mathcal{P} \rvert} \sum_{\mathbf{p}_i \in \mathcal{P}} \lVert \mathcal{W}(\mathbf{p}_i) - \mathcal{W}^{*}(\mathbf{p}_i) \rVert_2.
\end{equation}

\emph{3D Accuracy Strict} (AccS) and \emph{3D Accuracy Relaxed} (AccR) measure the fractions of points whose EPEs are below a EPE threshold or relative errors are below a relative error threshold. For AccS, the EPE threshold is $2.5$cm and the relative error threshold is $2.5\%$. For AccR, the EPE threshold is $5$cm and the relative error threshold is $5\%$. The relative error is computed as:
\begin{equation}
\mathtt{RE}(\mathbf{p}_i) = \frac{\lVert \mathcal{W}(\mathbf{p}_i) - \mathcal{W}^{*}(\mathbf{p}_i) \rVert_2}{\lVert \mathcal{W}^{*}(\mathbf{p}_i) - \mathbf{p}_i \rVert_2}.
\end{equation}
And the 3D accuracy is defined as:
\begin{align}
\mathtt{AccS} & \tight{=} \frac{1}{\lvert \mathcal{P} \rvert} \sum_{\mathbf{p}_i \in \mathcal{P}} \llbracket \mathtt{EPE}(\mathbf{p}_i) \tight{<} 2.5\text{cm} \tight{\lor} \mathtt{RE}(\mathbf{p}_i) \tight{<} 2.5\% \rrbracket,\\
\mathtt{AccR} & \tight{=} \frac{1}{\lvert \mathcal{P} \rvert} \sum_{\mathbf{p}_i \in \mathcal{P}} \llbracket \mathtt{EPE}(\mathbf{p}_i) \tight{<} 5\text{cm} \lor \mathtt{RE}(\mathbf{p}_i) \tight{<} 5\% \rrbracket,
\end{align}
where $\llbracket \cdot \rrbracket$ is the Inversion bracket.

\emph{Outlier Ratio} (OR) measures the fraction of points which are not successfully registered. Following~\cite{li2022non}, a point is regarded as a failure if its relative error is above $30\%$:
\begin{equation}
\mathtt{OR} = \frac{1}{\lvert \mathcal{P} \rvert} \sum_{\mathbf{p}_i \in \mathcal{P}} \llbracket \mathtt{RE}(\mathbf{p}_i) > 30\% \rrbracket
\end{equation}

\section{Additional Experiments}
\label{appx:additional-experiments}

\subsection{Evaluations on Low-Inlier-Ratio Cases}

To evaluate the performance in low-inlier-ratio scenarios, we add random outliers into the correspondences from GeoTransformer, making the final inlier ratio less than $30\%$. In \cref{table:results-low-ir}, PointDSC and PointCN fail to achieve reasonable registration results due to enormous outliers. In contrast, our method still achieves promising results, showing strong generality to low-inlier-ratio cases.

\subsection{Evaluations on Large-Deformation Cases}

Next, we investigate the performance when the deformations are large. As there is no off-the-shelf benchmarks with large deformations, we evaluate our method on the testing pairs whose mean residuals are above $15$cm on 4DMatch. In \cref{table:results-large-deformation}, our method significantly outperforms the baselines, demonstrating its efficacy under large deformations.

\subsection{Additional Ablation Studies}

\ptitle{Euclidean distance \vs geodesic distance.}
We first replace the distance metric in building deformation graph from Euclidean distance to geodesic distance. Each correspondence is assigned to its $k=6$ nearest neighbors in the geodesic space. Note that we still use Euclidean distance during N-ICP for fair comparison. As shown in \cref{table:results-ablation-supp}~(a), geodesic distance consistently degrades the performance. Compared to the Euclidean distance, the geodesic distance is less robust to occlusion as the points on the geodesic shortest path between two points can be missing. On the contrary, according to local rigidity, Euclidean distance is approximatedly preserved near each graph node, but is more robust and efficient.

\ptitle{Positional embedding.}
Next, we study the impact of the positional embedding used in the initial feature embedding in \cref{table:results-ablation-supp}~(b). We first ablate the the fourier positional encoding and use only the point coordinates. This model achieves similar results on 4DMatch and slightly worse results on 4DLoMatch. We then ablate the point coordinates and use only the fourier positional encoding. This model achieves better recall but worse precision, especially in low-overlap scenarios. And the model with the both terms achieve the best results.

\ptitle{Loss functions.}
We further study the efficacy of the loss functions in \cref{table:results-ablation-supp}~(c). We first ablate the feature consistency loss, which degrades the classification performance especially in low-overlap scenarios. Explicitly supervising the feature consistency between correspondences helps learn more discriminative features between inliers and outliers and thus contributes to better performance. Next we replace the binary focal loss with a binary cross-entropy loss, which significantly decreases the performance. As the putative correspondences are commonly extremely unbalanced, either predominated by inliers or outliers, cross-entropy loss hampers the convergency of the model.

\ptitle{Local spatial consistency.}
At last, we ablate the local spatial consistency in the self-attention. In \cref{table:results-ablation-supp}~(d), removing the local spatial consistency considerably decreases the performance, especially in low-overlap scenarios. We also note that this model surpasses PointCN and PointDSC, indicating the efficacy of our deformation graph-based design.


\begin{table}[t]
\scriptsize
\setlength{\tabcolsep}{2.5pt}
\centering
\begin{tabular}{l|cccc|cccc}
\toprule
 \multirow{2}{*}{Model} & \multicolumn{4}{c|}{4DMatch} & \multicolumn{4}{c}{4DLoMatch} \\
 & Prec & Recall & AccS & AccR & Prec & Recall & AccS & AccR \\
\midrule
GraphSCNet & \textbf{91.9} & 69.7 & \textbf{54.5} & \textbf{66.5} & \textbf{81.8} & 70.6 & \textbf{26.9} & \textbf{38.6} \\
PointDSC & 55.3 & 81.5 & 5.8 & 12.7 & 50.6 & 74.6 & 4.8 & 10.5 \\
PointCN & 44.8 & \textbf{85.1} & 3.1 & 10.4 & 42.3 & \textbf{74.8} & 3.4 & 8.9 \\
w/o outlier rejection & 29.3 & 100.0 & 0.5 & 1.9 & 25.7 & 100.0 & 1.0 & 2.9 \\
\bottomrule
\end{tabular}
\caption{
Evaluations on 4DMatch and 4DLoMatch with low inlier ratios.
}
\label{table:results-low-ir}
\end{table}


\begin{table}[t]
\scriptsize
\setlength{\tabcolsep}{2.5pt}
\centering
\begin{tabular}{l|cccc|cccc}
\toprule
 \multirow{2}{*}{Model} & \multicolumn{4}{c|}{4DMatch} & \multicolumn{4}{c}{4DLoMatch} \\
 & Prec & Recall & AccS & AccR & Prec & Recall & AccS & AccR \\
\midrule
GraphSCNet & \textbf{89.1} & \textbf{93.8} & \textbf{57.1} & \textbf{71.5} & \textbf{77.8} & \textbf{78.4} & \textbf{28.6} & \textbf{42.7} \\
PointDSC & 83.3 & 90.5 & 53.6 & 68.7 & 66.4 & 76.7 & 25.7 & 40.1 \\
PointCN & 80.0 & 87.5 & 51.1 & 67.0 & 62.4 & 75.5 & 23.3 & 38.4 \\
w/o outlier rejection & 76.4 & 100.0 & 51.7 & 68.3 & 56.2 & 100.0 & 23.5 & 38.9 \\
\bottomrule
\end{tabular}
\caption{
Evaluations on 4DMatch and 4DLoMatch with large deformations.
}
\label{table:results-large-deformation}
\end{table}


\begin{table}[t]
\scriptsize
\setlength{\tabcolsep}{3pt}
\centering
\begin{tabular}{l|cccc|cccc}
\toprule
\multirow{2}{*}{Model} & \multicolumn{4}{c|}{4DMatch} & \multicolumn{4}{c}{4DLoMatch} \\
 & Prec & Recall & AccS & AccR & Prec & Recall & AccS & AccR \\
\midrule
(a.1) Euclidean & \textbf{92.2} & \textbf{96.9} & \textbf{72.3} & \textbf{84.4} & \textbf{82.6} & \textbf{86.8} & \textbf{41.0} & \textbf{58.3} \\
(a.2) Geodesic & 91.2 & 96.4 & 71.1 & 83.5 & 80.7 & 85.6 & 39.8 & 57.4 \\
\midrule
(b.1) XYZ+Fourier & \textbf{92.2} & \underline{96.9} & \textbf{72.3} & \textbf{84.4} & \textbf{82.6} & \underline{86.8} & \textbf{41.0} & \textbf{58.3} \\
(b.2) XYZ & \textbf{92.2} & 96.6 & \textbf{72.3} & \textbf{84.4} & \underline{82.1} & 86.1 & \underline{40.8} & \underline{58.0} \\
(b.3) Fourier & 91.3 & \textbf{97.2} & 71.9 & 84.2 & 79.8 & \textbf{87.1} & 40.2 & 57.5 \\
\midrule
(c.1) w/ FL w/ CL* & \textbf{92.2} & \textbf{96.9} & \textbf{72.3} & \textbf{84.4} & \textbf{82.6} & \textbf{86.8} & \textbf{41.0} & \textbf{58.3} \\
(c.2) w/ FL w/o CL & \underline{90.5} & \underline{96.4} & \underline{71.3} & \underline{83.7} & \underline{77.5} & \underline{84.4} & \underline{38.5} & \underline{55.8} \\
(c.3) w/ BCE w/o CL & 79.5 & 93.1 & 64.3 & 78.2 & 55.4 & 75.4 & 28.2 & 44.1 \\
\midrule
(d.1) w/ local SC & \textbf{92.2} & \textbf{96.9} & \textbf{72.3} & \textbf{84.4} & \textbf{82.6} & \textbf{86.8} & \textbf{41.0} & \textbf{58.3} \\
(d.2) w/o local SC & 90.5 & 97.0 & 71.3 & 83.6 & 78.9 & 88.7 & 39.6 & 56.9 \\
\bottomrule
\end{tabular}
\caption{
Additional Ablation studies on 4DMatch and 4DLoMatch.
Asterisk (*) indicates the default settings in our method.
\textbf{FL}: focal loss.
\textbf{CL}: consistency loss.
\textbf{BCE}: binary cross-entropy loss.
\textbf{Boldfaced} numbers highlight the best and the second best are \underline{underlined}.
}
\label{table:results-ablation-supp}
\end{table}

\subsection{Qualitative Results}


\begin{figure}[t]
  \centering
  \begin{overpic}[width=1.0\linewidth]{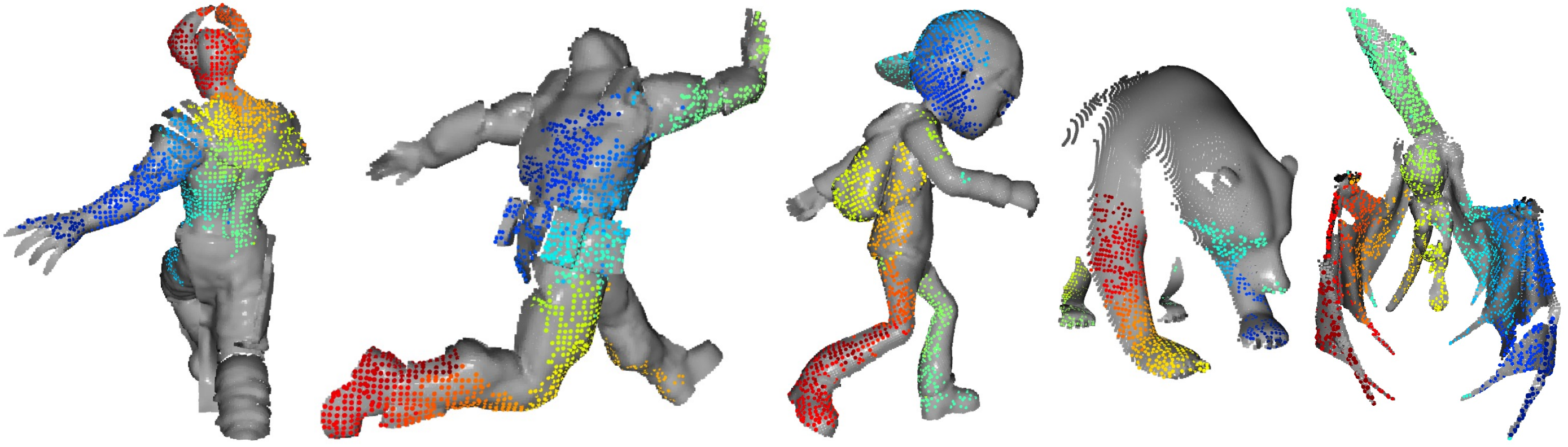}
  \end{overpic}
  \caption{Feature distribution of inliers.}
  \label{fig:distribution}
\end{figure}

We visualize the features of the detected true inliers by t-SNE. In \cref{fig:distribution}, the inliers in different parts have different features, while the spatially-near ones also lie closely in the feature space. These results indicate that our method effectively learns the local motions in different parts.

\begin{figure*}[t]
  \centering
  \begin{overpic}[width=1.0\linewidth]{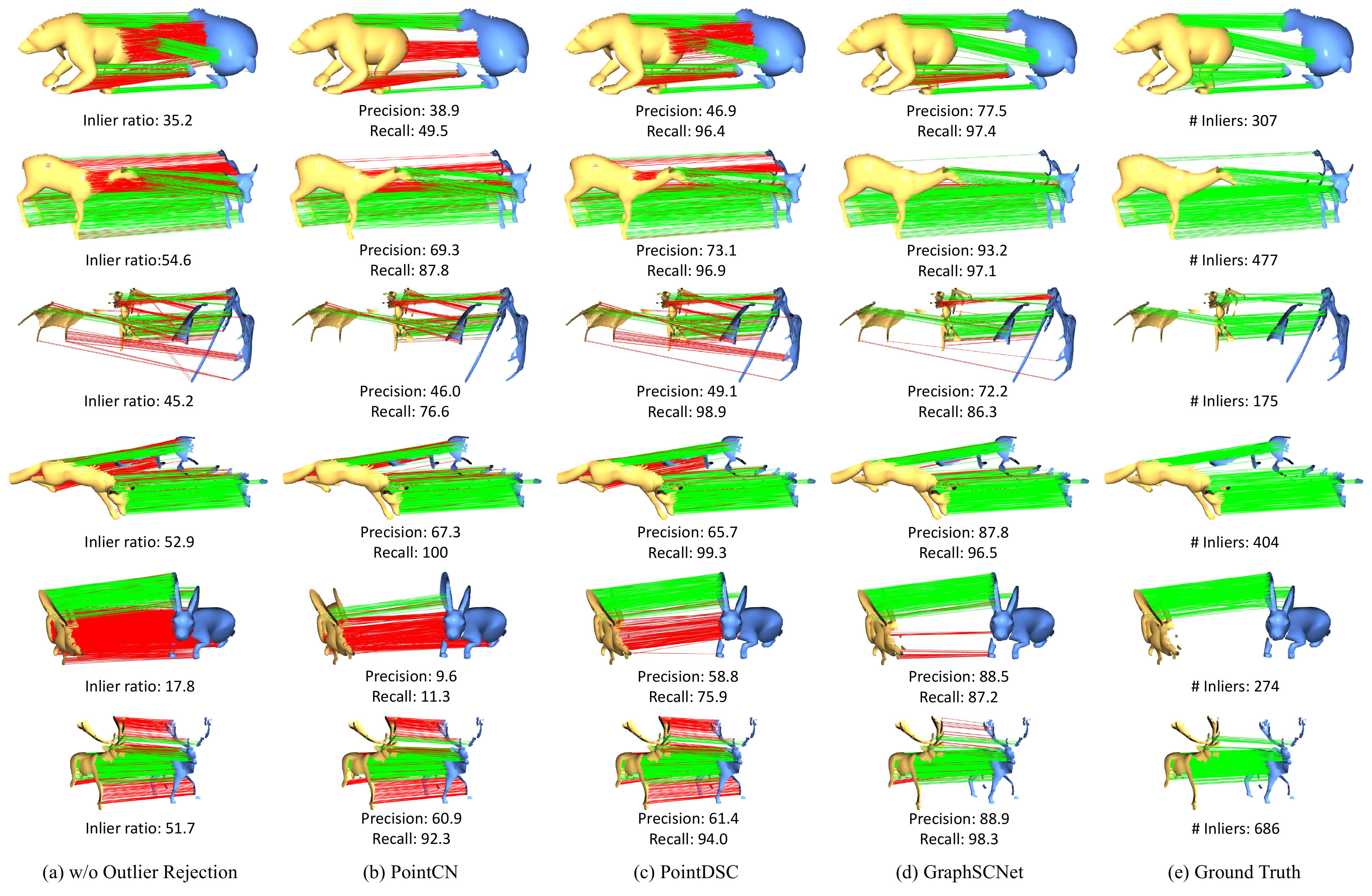}
  \end{overpic}
  \caption{Comparison of different methods on 4DMatch and 4DLoMatch.}
  \label{fig:gallery-4dmatch-supp}
\end{figure*}

\begin{figure*}[t]
  \centering
  \begin{overpic}[width=1.0\linewidth]{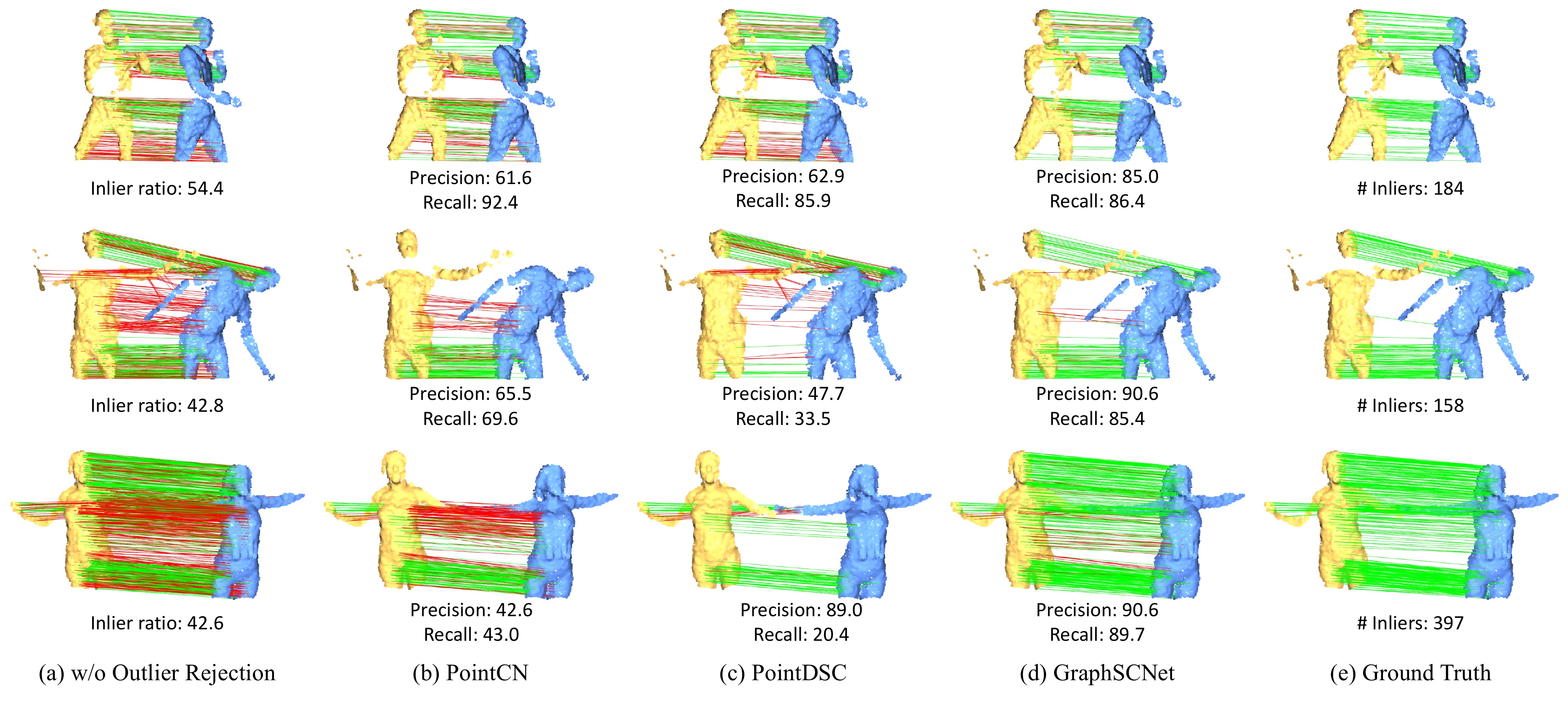}
  \end{overpic}
  \caption{
  Comparison of different methods on DeepDeform.
  }
  \label{fig:gallery-deepdeform}
\end{figure*}

\begin{figure*}[t]
  \centering
  \begin{overpic}[width=1.0\linewidth]{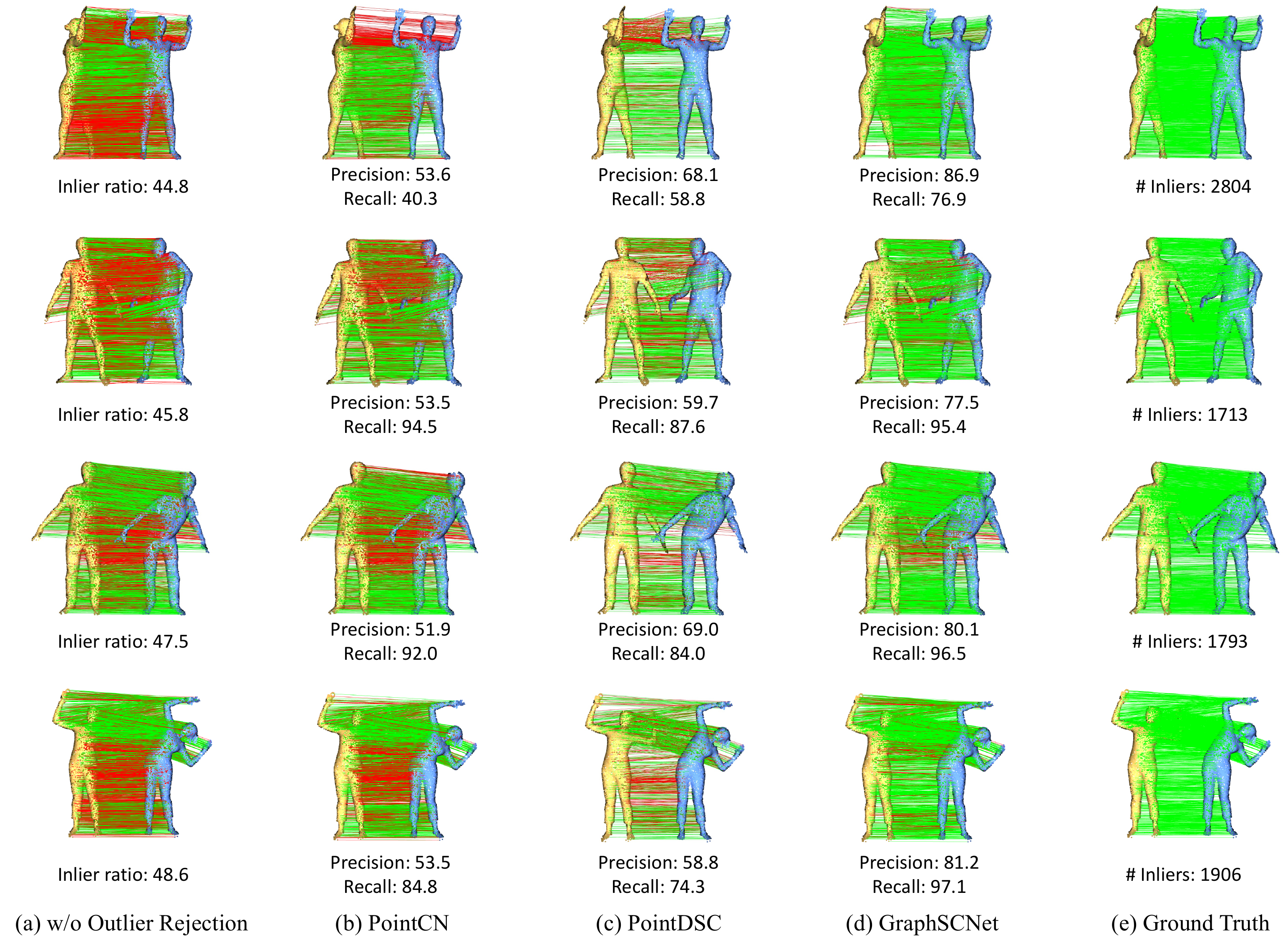}
  \end{overpic}
  \caption{
  Comparison of different methods on DeepDeform.
  }
  \label{fig:gallery-cape-supp}
\end{figure*}

We then provide more qualitative comparisons of the filtered correspondences on 4DMatch (\cref{fig:gallery-4dmatch-supp}), CAPE (\cref{fig:gallery-cape-supp}) and DeepDeform (\cref{fig:gallery-deepdeform}).
Benefitting to the powerful local spatial consistency, \ours{} removes more outlier correspondences and achieves better inlier ratio (precision) than the baseline methods, especially under large deformations. Moreover, albeit achieving promising precision, PointDSC fails to preserve sufficient inliers. On the contrary, our method achieves both high precision and high recall, indicating it can effectively reject most outliers while better preserving inliers.

\section{Deformation Estimation}
\label{appx:deformation-estimation}

Given the source point cloud $\mathcal{P}$, the target point cloud $\mathcal{Q}$, and the correspondences $\mathcal{C} = \{(\mathbf{x}_i, \mathbf{y}_i) \mid \mathbf{x}_i \in \mathcal{P}, \mathbf{y}_i \in \mathcal{Q} \}$ between them, we adopt embedded deformation~\cite{sumner2007embedded} to formulate the warping function. It parameterizes the deformation on the deformation graph $\mathcal{G}=\{\mathcal{V}, \mathcal{E}\}$.
The graph nodes $\mathcal{V}$ are sampled from the \emph{source} point cloud with uniform furthest point sampling and the node coverage is $\sigma_{g} = 0.08\text{m}$. Each point $\mathbf{p}_i$ in the source point cloud is assigned to its $k_g = 6$ nearest nodes $\mathcal{K}_i$. Two nodes are connected by an undirected edge if they share a common point. Each node $\mathbf{v}_j$ is associated with a local rigid transformation $\{ \mathbf{R}_j, \mathbf{t}_j \}$. And the warping function $\mathcal{W}$ is then approximated as:
\begin{equation}
\mathcal{W}(\mathbf{p}_i) = \sum_{\mathbf{v}_j \in \mathcal{V}} \alpha_{i,j} \big(\mathbf{R}_j (\mathbf{p}_i - \mathbf{v}_j) + \mathbf{t}_j + \mathbf{v}_j\big),\notag
\end{equation}
where $\alpha_{i, j}$ is the skinning factor as defined in~\cite{newcombe2015dynamicfusion}:
\begin{equation}
\alpha_{i, j} = \llbracket \mathbf{v}_j \in \mathcal{K}_i \rrbracket \cdot \frac{\exp(-\lVert \mathbf{p}_i - \mathbf{v}_j \rVert^2 / (2 \sigma_n^2))}{\sum_{\mathbf{v}_k \in \mathcal{K}_i} \exp(-\lVert \mathbf{p}_i - \mathbf{v}_k \rVert^2 / (2 \sigma_n^2))},\notag
\end{equation}
where $\llbracket \cdot \rrbracket$ is the Iverson bracket.
We then solve for $\mathcal{W}$ by minimizing the following objective function:
\begin{equation}
E = \lambda_c E_{\text{corr}} + \lambda_r E_{\text{reg}},\notag
\end{equation}
where $E_{\text{corr}}$ is the mean squared residual between the correspondences and $E_{\text{reg}}$ is an as-rigid-as-possible~\cite{igarashi2005rigid} regularization term to constrain the smoothness of deformations:
\begin{equation}
\begin{aligned}
E_{\text{corr}} & = \sum_{(\mathbf{x}_i, \mathbf{y}_i) \in \mathcal{C}} \lVert \mathcal{W}(\mathbf{x}_i) - \mathbf{y}_i \rVert_{2}^{2},\\
E_{\text{reg}} & = \sum_{(\mathbf{v}_u, \mathbf{v}_v) \in \mathcal{E}} \lVert \mathbf{R}_{u}(\mathbf{v}_{v} - \mathbf{v}_{u}) + \mathbf{v}_{u} + \mathbf{t}_{u} - (\mathbf{v}_{v} + \mathbf{t}_{v}) \rVert_{2}^{2}.
\end{aligned}
\notag
\end{equation}
The weights to balance the two terms are set to $\lambda_c = 25$ and $\lambda_r = 1$, respectively.

This problem can be efficiently solved by Non-rigid ICP (N-ICP) algorithm~\cite{li2008global,sumner2007embedded}.
Following~\cite{bozic2020neural,li2022lepard}, we update the associated rigid transformations incrementally:
\begin{equation}
\begin{aligned}
\mathbf{R}^{(t)}_j & = \Delta\mathbf{R}^{(t)}_j \cdot \mathbf{R}^{(t-1)}_j,\\
\mathbf{t}^{(t)}_j & = \mathbf{t}^{(t-1)}_j + \Delta\mathbf{t}^{(t)}_j,
\end{aligned}
\notag
\end{equation}
where $\mathbf{R}^{(0)}_j = \mathbf{I}$ and $\mathbf{t}^{(0)}_j = \mathbf{0}$. For simplicity, we omit the superscript $(t)$ in the following text.
The residual rotations are formulated in the axis-angle representation $\Delta\mathbf{R}_j \tight{=} \exp(\boldsymbol{\omega}^{\land}_j)$, where $\exp(\cdot)$ is the exponential map function and $(\cdot)^{\land}$ computes the skew-symmetric matrix of a $3$-d vector.
We then solve for $\{ \boldsymbol{\omega}_j, \Delta\mathbf{t}_j \}$ with Gauss-Newton algorithm. The residual terms are computed as:
\begin{equation}
\begin{aligned}
\mathbf{r}^{i}_{\text{corr}} & \tight{=} \sqrt{\lambda_{c}} \bigl( \mathcal{W}(\mathbf{x}_i) - \mathbf{y}_i \bigr),\\
\mathbf{r}^{i}_{\text{reg}} & \tight{=} \sqrt{\lambda_{r}} \bigl( \mathbf{R}_{u}(\mathbf{v}_{v} - \mathbf{v}_{u}) + \mathbf{v}_{u} + \mathbf{t}_{u} - (\mathbf{v}_{v} + \mathbf{t}_{v}) \bigr).
\end{aligned}
\notag
\end{equation}
where $c_i = (\mathbf{x}_i, \mathbf{y}_i) \in \mathcal{C}$ and $e_i = (\mathbf{v}_u, \mathbf{v}_v) \in \mathcal{E}$.

Next, we compute the partial derivatives of $\{ \boldsymbol{\omega}_j, \Delta\mathbf{t}_j \}$.
As $\boldsymbol{\omega}_j$ is a residual rotation, it is expected to near $\mathbf{0}$ and thus we approximate its partial derivatives with those at $\mathbf{0}$:
\begin{equation}
\begin{aligned}
\frac{\partial \mathcal{W}(\mathbf{p}_i)}{\partial \boldsymbol{\omega}_j} & \approx \frac{\partial \mathcal{W}(\mathbf{p}_i)}{\partial \boldsymbol{\omega}_j}\bigg\arrowvert_{\mathbf{0}} = -\alpha_{i, j} \bigl(\mathbf{R}^{(t-1)}_j(\mathbf{x}_i - \mathbf{v}_j)\bigr)^{\land},\\
\frac{\partial \mathcal{W}(\mathbf{p}_i)}{\partial \Delta\mathbf{t}_j} & = \alpha_{i, j} \mathbf{I}.
\end{aligned}
\notag
\end{equation}
To this end, the partial derivatives are computed as:
\begin{equation}
\begin{aligned}
\frac{\partial \mathbf{r}^{i}_{\text{corr}}}{\partial \boldsymbol{\omega}_j} & = -\sqrt{\lambda_c} \alpha_{i, j} \bigl(\mathbf{R}^{(t-1)}_j(\mathbf{x}_i - \mathbf{v}_j)\bigr)^{\land},\\
\frac{\partial \mathbf{r}^{i}_{\text{corr}}}{\partial \Delta\mathbf{t}_j} & = \sqrt{\lambda_c} \alpha_{i, j} \mathbf{I},\\
\frac{\partial \mathbf{r}^{i}_{\text{reg}}}{\partial \boldsymbol{\omega}_u} & = -\sqrt{\lambda_r} \bigl(\mathbf{R}^{(t-1)}_u(\mathbf{v}_v - \mathbf{v}_u)\bigr)^{\land},\\
\frac{\partial \mathbf{r}^{i}_{\text{reg}}}{\partial \Delta\mathbf{t}_u} & = \sqrt{\lambda_r} \mathbf{I},\\
\frac{\partial \mathbf{r}^{i}_{\text{reg}}}{\partial \Delta\mathbf{t}_v} & = -\sqrt{\lambda_r} \mathbf{I}.
\end{aligned}
\notag
\end{equation}
We denote the collection of the residual terms as:
\begin{equation}
\mathbf{r} = [(\mathbf{r}^{1}_{\text{corr}})^T, ..., (\mathbf{r}^{\lvert \mathcal{C} \rvert}_{\text{corr}})^T, (\mathbf{r}^{1}_{\text{reg}})^T, ..., (\mathbf{r}^{\lvert \mathcal{E} \rvert}_{\text{reg}})^T]^T \in \mathbb{R}^{3 \lvert \mathcal{C} \rvert + 3 \lvert \mathcal{E} \rvert},
\notag
\end{equation}
the collections of variables $\{\boldsymbol{\omega}_j, \Delta\mathbf{t}_j \}$ as:
\begin{equation}
\Delta\mathbf{T} = [\boldsymbol{\omega}^T_1, ..., \boldsymbol{\omega}^T_{\lvert \mathcal{V} \rvert}, \Delta\mathbf{t}^T_1, ..., \Delta\mathbf{t}^T_{\lvert \mathcal{V} \rvert}]^T \in \mathbb{R}^{6 \lvert \mathcal{V} \rvert},
\notag
\end{equation}
and the Jaccobian matrix between $\mathbf{r}$ and $\Delta\mathbf{T}$ is denoted as $\mathbf{J} \tight{\in} \mathbb{R}^{(3 \lvert \mathcal{C} \rvert + 3 \lvert \mathcal{E} \rvert) \times (6 \lvert \mathcal{V} \rvert)}$ following the computation of derivatives above. $\Delta\mathbf{T}$ can then be computed by solving the linear system:
\begin{equation}
(\mathbf{J}^T\mathbf{J} + \lambda_m \mathbf{I})\Delta\mathbf{T} = \mathbf{J}^T \mathbf{r}.
\notag
\end{equation}
where $\lambda_m=0.01$ is the Marquardt factor.

\section{Limitations}
\label{appx:limitations}

Our method could have the following two potential limitations. First, our method serves as a post outlier rejection step after the correspondence extractor. To this end, our method is able to make given correspondences as clean as possible, but cannot infer new correspondences and improve the coverage of the correspondences on point clouds. Second, our method is based on deformation graph and local rigidity of deformations, so it could have difficulty in modeling sudden changes of geometric structures. We would leave these for future work.


{\small
\bibliographystyle{ieee_fullname}
\bibliography{graphsc}
}

\end{document}